\documentclass[10pt,journal]{IEEEtran}

\ifCLASSOPTIONcompsoc
  \usepackage[nocompress,noadjust]{cite}
\else
  \usepackage{cite}
\fi

\ifCLASSINFOpdf
  \usepackage[pdftex]{graphicx}
  \graphicspath{{./pdf/}{./jpeg/}{./graphics/}{./ME/Figs/}{./photos/}}
\else
\fi

\usepackage{amsmath}
\usepackage{lipsum}
\usepackage{mathtools}
\usepackage{cuted}
\usepackage{comment}
\ifCLASSOPTIONcompsoc
  \usepackage[caption=false,font=footnotesize,labelfont=sf,textfont=sf]{subfig}
\else
  \usepackage[caption=false,font=footnotesize]{subfig}
\fi
\usepackage{ragged2e}

\usepackage{tikz}
\usetikzlibrary{shapes,arrows,scopes,positioning,calc,math,patterns,decorations.pathmorphing,decorations.markings,decorations.pathreplacing,spy,angles,quotes}
\usepackage{ctable}
\makeatletter
\def\thickhline{%
  \noalign{\ifnum0=`}\fi\hrule \@height \thickarrayrulewidth \futurelet
   \reserved@a\@xthickhline}
\def\@xthickhline{\ifx\reserved@a\thickhline
               \vskip\doublerulesep
               \vskip-\thickarrayrulewidth
             \fi
      \ifnum0=`{\fi}}

\usepackage{array}

\pretocmd\@bibitem{\color{black}\csname keycolor#1\endcsname}{}{\fail}
\newcommand\citecolor[1]{\@namedef{keycolor#1}{\color{red}}}
\makeatother

\newlength{\thickarrayrulewidth}
\usepackage{multirow}
\usepackage{makecell}
\usepackage{mcode}

\usepackage[inline]{asymptote}

\usepackage{siunitx}
\usepackage{nicefrac}

\let\oldfrac\frac
\makeatletter
\newcommand{\groupit}[1]{(#1)}
\newcommand{\nogroupit}[1]{#1}
\renewcommand{\frac}[2]{%
  \setbox\z@\hbox{$#1$}
  \setbox\tw@\hbox{$#2$}
  \ifdim\wd\z@>1em \let\groupornot@i\groupit\else\let\groupornot@i\nogroupit\fi
  \ifdim\wd\tw@>1em \let\groupornot@ii\groupit\else\let\groupornot@ii\nogroupit\fi
  \mathchoice
    {\oldfrac{#1}{#2}}
    {\groupornot@i{#1}/\groupornot@ii{#2}}
    {\groupornot@i{#1}/\groupornot@ii{#2}}
    {\groupornot@i{#1}/\groupornot@ii{#2}}
}
\usepackage{xcolor}
\usepackage[normalem]{ulem}

\usepackage{soul}
\usepackage{lscape}
\usepackage[utf8]{inputenc}
\usepackage{float}
\usepackage{rotating}
\usepackage{pifont}
\usepackage{enumitem}
\usepackage{hyperref}
\usepackage[bbgreekl]{mathbbol}
\usepackage{cuted}
\usepackage[linesnumbered,lined,boxed]{algorithm2e}

\usepackage{tabularray}

\begin{document}
\title{Towards A General-Purpose Motion Planning for Autonomous Vehicles Using Fluid Dynamics}

\author{MReza~Alipour~Sormoli$^{1}$, Konstantinos Koufos$^{1}$,
Mehrdad~Dianati$^{1,2}$,~\IEEEmembership{Senior~Member,~IEEE,}
and Roger Woodman$^{1}$%
\thanks{$^{1}$ WMG, University of Warwick, Coventry, CV4 7AL, UK. {\tt\footnotesize mreza.alipour@warwick.ac.uk}
}
\thanks{$^{2}$ School of Electronics, Electrical Engineering and Computer Science, Queen’s University of Belfast, UK. {\tt\footnotesize m.dianati@qub.ac.uk}}
}

\markboth{-}{}%

\maketitle

\begin{abstract}

General-purpose motion planners for automated/autonomous vehicles promise to handle the task of motion planning (including tactical decision-making and trajectory generation)  for various automated driving functions (ADF) in a diverse range of operational design domains (ODDs). The challenges of designing a general-purpose motion planner arise from several factors: a) A plethora of scenarios with different semantic information in each driving scene should be addressed, b) a strong coupling between long-term decision-making and short-term trajectory generation shall be taken into account, c) the nonholonomic constraints of the vehicle dynamics must be considered, and d) the motion planner must be computationally efficient to run in real-time.  The existing methods in the literature are either limited to specific scenarios (logic-based) or are data-driven (learning-based) and therefore lack explainability, which is important for safety-critical automated driving systems (ADS). This paper proposes a novel general-purpose motion planning solution for ADS inspired by the theory of fluid mechanics. 
A computationally efficient technique, i.e., the lattice Boltzmann method, is then adopted to generate a spatiotemporal vector field, which in accordance with the nonholonomic dynamic model of the Ego vehicle is employed to generate feasible candidate trajectories. The trajectory optimising ride quality, efficiency and safety is finally selected to calculate the imminent control signals, i.e., throttle/brake and steering angle. The performance of the proposed approach is evaluated by simulations in highway driving, on-ramp merging, and intersection crossing scenarios, and it is found to outperform traditional motion planning solutions based on model predictive control (MPC). 

\end{abstract}

\begin{IEEEkeywords}
Autonomous/automated driving systems, autonomous vehicles, fluid dynamics, lattice Boltzmann method, motion planning, trajectory generation.
\end{IEEEkeywords}

\IEEEpeerreviewmaketitle

\section{Introduction}
\label{sec: introduction}
\IEEEPARstart{D}{esigning} highly autonomous/automated driving systems (ADS) is currently an active area of research in academia and the automotive industry, motivated by the foreseen benefits in safety, fuel efficiency, ride quality, and accessibility of future transport systems, to name a few. In a nutshell, an ADS implements a perception, planning (mission, motion, and trajectory generation) and actuator loop in real-time. In complex environments, where a great variety of actors with different behaviours and intentions share the road, an autonomous vehicle must be able to successfully react in real-time based on the perception of the surrounding environment. In addition, the nonholonomic dynamic constraint of a four-wheeled vehicle shall be considered in the motion planning and control of the vehicle. Developing motion planning algorithms that can handle the abovementioned challenges under various driving scenarios and operational design domains (ODDs) remains challenging. 

Motion planning consists of two main functions: Tactical decision-making and trajectory generation/planning. The former refers to the decisions and manoeuvres performed by the vehicle, such as lane change, highway merging and overtaking. The latter focuses on fine-grained motion features, such as the velocity and yaw rate signals for the next few seconds. 
Tactical decision-making and trajectory generation can be combined either explicitly or implicitly for motion planning. In explicit methods, the decision-making and trajectory generation processes are distinct and interact in a hierarchical framework. The decision-making process is updated at a lower frequency of $1$-$5$~Hz based upon requests received from the trajectory generation process that needs to update the generated trajectory at a higher rate of $10$-$20$~Hz~\cite{claussmann2019review}. If a feasible trajectory cannot be found for the selected manoeuvre, the decision-making layer must provide an alternative manoeuvre from a set of predefined manoeuvres. As a result, despite their computational efficiency, explicit motion planners are likely to miss the optimal trajectory. Furthermore,  adapting them to a different scenario or ODD may require heavy modifications and tuning of the underlying algorithms for decision-making and trajectory generation~\cite{lim2019hybrid}. Therefore, explicit motion planners are also likely to perform poorly during rare unknown events. 

In implicit motion planning, the tactical decision-making and trajectory generation functions are used to formulate a single optimisation problem, hence, not compromising feasibility and optimality. Most of the existing implicit motion planners adopt a data-driven approach because of their ability to adapt to various driving scenarios and conditions by retraining the system. However, data-driven methods require large datasets for training and they are often unexplainable, which is a crucial requirement in safety-critical systems such as ADS. As an alternative approach, various optimisation techniques, such as spatiotemporal state lattice search~\cite{mcnaughton2011motion}, and model predictive controllers (MPCs)~\cite{hang2020human,dixit2019trajectory} can be used to solve implicit motion planning problems. These algorithms not only require some modifications to adapt to different driving scenarios and ODDs, but they also suffer from elevated computational demands, and susceptibility to local minima with non-convex cost functions.

Motivated by the shortcomings of the existing approaches in the literature to solving implicit motion planning problems, this paper proposes a novel approach based on the theory of fluid flow simulation in a confined domain with the following features: 1) {\it{Generality in terms of the underlying ADF}} by formulating the driving context, including diverse semantic elements such as road boundaries, lane markings and other road users into a spatiotemporal fluid flow boundary value problem. 2) {\it{Computational efficiency}} by using the lattice Boltzmann method (LBM) for simulating the fluid flow motion via parallel computation at a low computational demand in contemporary hardware. 3) {\it{Optimality}} by defining a cost function that factors in efficiency, safety and comfort, and selecting the optimal trajectory accordingly. 4) {\it{Practicality}} by considering the nonlinear dynamics and nonholonomic constraints of the vehicle, therefore addressing the feasibility concerns of the planned motion. 

\begin{figure*}[t]
\centering
\includegraphics[width=.9\linewidth]{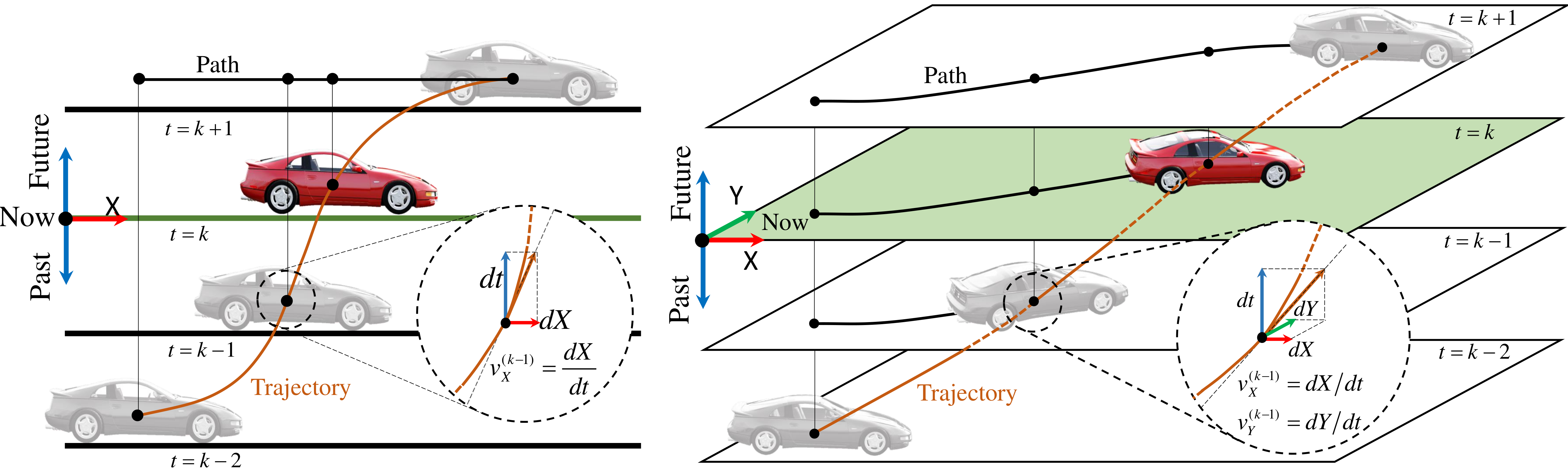}
\caption{Spatiotemporal trajectory (brown curves) of the EV moving on a) 1D straight line and b) 2D plane with the vertical axis representing time. The 1D example helps visualise that the speed of the vehicle at each time step is given by the gradient of the spatiotemporal trajectory.}
\label{fig: spatiotemporal traj}
\centering
\end{figure*}

 In our approach, the solution space for the candidate trajectories is interpreted as a 3D structured spatiotemporal domain in which fluid flows, e.g., moving obstacles in the 2D plane become 3D static obstacles deviating fluid motion. As the fluid particles flow on an optimum trajectory, i.e., continuous trajectory with minimum energy consumption transporting a unit of fluid quantity~\cite{sochi2014solving}, the main idea behind the proposed method is to use fluid motion streamlines for sampling trajectories in the spatiotemporal domain, see Fig.~\ref{fig: spatiotemporal traj} for example illustrations. The temporal and spatial features of the calculated streamlines are not decoupled in our method; in other words, the manoeuvre and the velocity profile are jointly calculated. In this way, the feasibility and optimality of the discovered trajectories are not compromised. Unlike explicit motion planning techniques we do not predefine the type of the manoeuvre, but we encode diverse road elements and other semantic information, such as lane markings, traffic rules, and moving/static obstacles into a fluid domain with boundary conditions. For instance, in a merging scenario, the fluid-inspired motion planner continuously detects the appropriate gap for the Ego vehicle (EV) to merge in and the time to start the lane change without deciding in advance on the driving behaviour. The EV can dynamically change the selected gap if another vehicle reacts in a way that impedes the EV manoeuvre. Hence, the implementation of the motion planning algorithm doesn't need to be changed for different ADFs, unlike optimisation-based methods, which are mostly scenario-specific. Finally, the proposed approach in this paper can be easily parallelised to improve its computational speed. 

The remainder of this paper is organized in the following sections: After reviewing related studies in Section~\ref{sec: related work}, a background on vehicle dynamics is given in Section~\ref{sec: problem definition}. In Section~\ref{sec: proposed mp}, the proposed method using fluid dynamics is described in detail and in Section~\ref{sec: evaluation} its performance evaluation is illustrated under various driving scenarios. This section also provides a comparative analysis with state-of-the-art methods is also discussed. Key highlights are concluded in Section~\ref{sec: conclusion}.

\begin{figure}[t]
\centering
\includegraphics[width=1\linewidth]{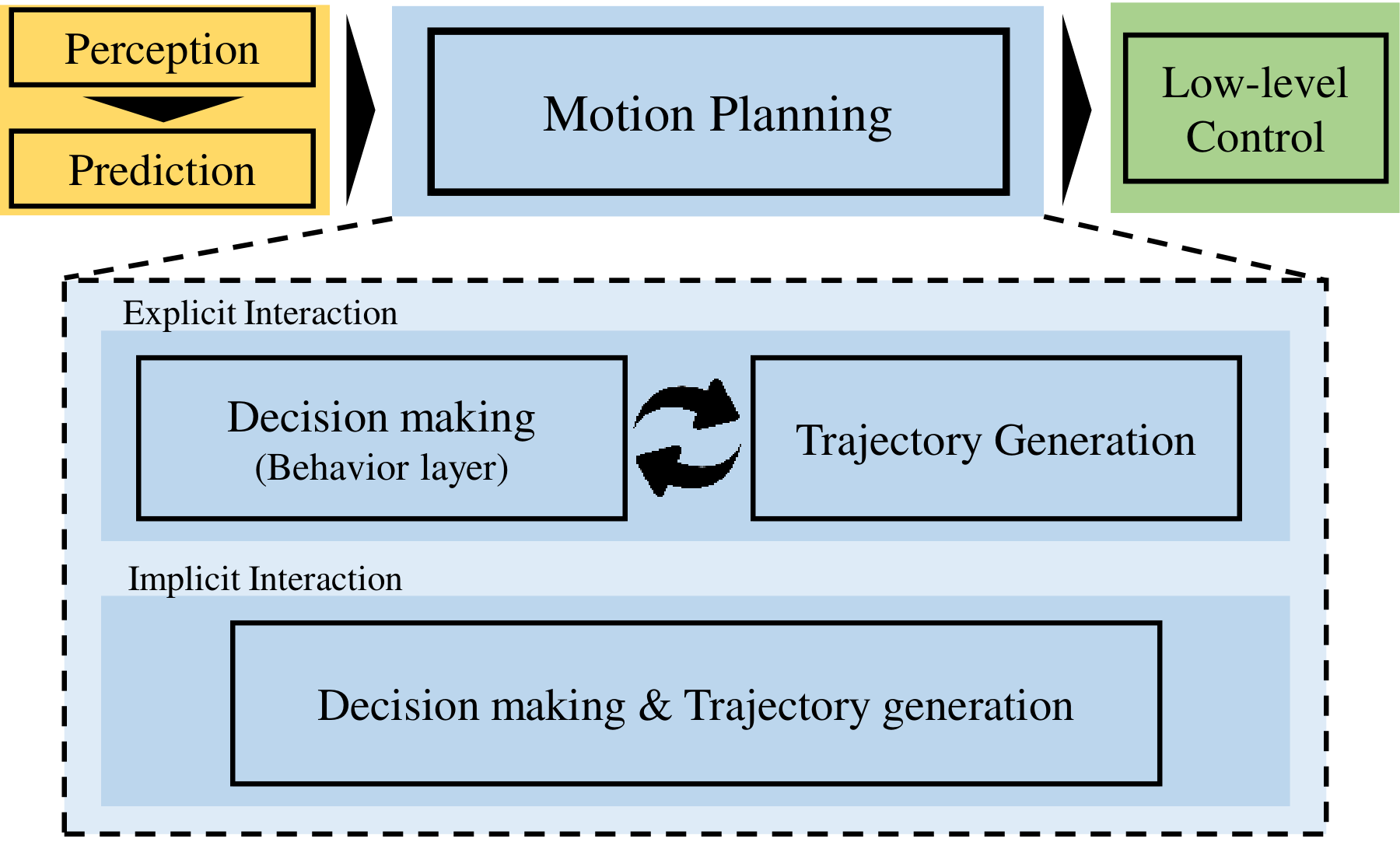}
\caption{Classification of motion planning design approaches within the general system diagram of an ADS. The motion planner is fed with the output of the perception module and the predictions of other road users' intentions before producing the signals for the low-level control of the Ego vehicle.}
\label{fig: system diagram}
\centering
\end{figure}

\section{Related Work}
\label{sec: related work}
Next, we review key studies on motion planning for ADS following the breakdown of motion planners into 
\textit{explicit} and \textit{implicit} depending on the type of interaction between decision-making and trajectory generation, see Fig.~\ref{fig: system diagram} for an overview. 

\subsection{Explicit motion planning methods}
In motion planning, the decision-making layer considers the high-level abstracted elements of the EV behaviour in a long time horizon (far future), and the trajectory generation process handles the detailed dynamic model of the vehicle and hard constraints in the imminent future. The authors in~\cite{nilsson2016if} employed a simple rule-based framework to decide whether a lane change is desirable in highway driving scenarios, and if so, they planned for the lateral/longitudinal trajectory to fit within the selected inter-vehicle traffic gap. In order to avoid defining numerous rules and conditions, a finite state machine was developed in~\cite{bae2020finite, wang2018predictive} to determine the selection rule among the predefined states. To cover more driving scenarios, lane-level motion planning methods were also introduced. Specifically, the study in~\cite{fan2018baidu} used a reference route for each lane. It first generated the paths considering all obstacles and traffic regulations and after that calculated smooth velocity profiles (in parallel) for each lane. Finally, the optimal trajectory in terms of speed profile smoothness and obstacle avoidance is selected. In another study~\cite{zhang2021unified}, the authors designed an integrated motion planning for highway scenarios using a 3D spatiotemporal voxel to model the coupling between lateral and longitudinal motion, ignoring, however, the non-holonomic constraints of the vehicle. In summary, even though the explicit interaction approach facilitates the motion planning design procedure, the final algorithms and the selected optimisation criteria are often tailored to specific driving scenarios/situations.

\subsection{Implicit motion planning methods}
The methods that combine decision-making with trajectory generation can tackle the motion planning problem including rare unforeseen situations. For instance, data-driven methods can uncover complex relationships between various elements in the driving scene, such as traffic regulations, driving area, positions/speeds of surrounding vehicles (SVs) and the desired manoeuvre or speed profile for the EV. Provided that the available training data is sufficient, data-driven methods can cope with various driving situations and ODDs.

Several studies have already focused on imitating the human driving style for motion planning using artificial neural networks. The authors in~\cite{li2018humanlike} introduced a fully connected neural network trained in a game engine environment and took an abstraction of the driving scene (e.g., the locations of SVs) as input to calculate the EV's steering angle and speed. 
Similarly, the study in~\cite{lefevre2015learning} used real driving data to train a neural network to perform car-following in combination with robust MPC. The confidence values of the learning model are fed into the MPC to decide whether or not the output of the neural network shall be trusted, empowering the motion planner with the capability to handle driving situations that are not present in the training dataset. Reinforcement learning was also used to learn the best policy (driving actions) by exploring the surrounding environment~\cite{shalev2016safe,shalev2017formal}. However, defining the reward function in ADS  leveraging RL is not straightforward and requires manual tuning or trial and error. In response to that challenge, maximum entropy inverse reinforcement learning was employed to derive the underlying reward function utilizing expert driver knowledge~\cite{rosbach2019driving,arora2021survey}. As expected, data-driven methods need further inspection to interpret their underlying logic for gaining trust and debugging (explainability) in the case of failure. 

Heuristic approaches that combine machine learning and traditional (rule-based or optimisation-based) techniques for motion planning can also handle tactical decision-making and low-level trajectory generation at once~\cite {mcnaughton2011motion}. In that case, optimality or real-time processing capabilities with contemporary hardware have to be inevitably sacrificed. For instance, although the conformal lattice structure considers the road geometry, that does not change based on the distribution of dynamic objects in the driving scene. On the one hand, that leads to a sparse search space with constant lattice size~\cite{mcnaughton2011motion, ziegler2009spatiotemporal}. On the other hand, excluding the occupied spatiotemporal domain (by moving/static objects) provides a higher-resolution lattice with the same number of edges at the cost of added complexity to construct a dynamic lattice.  Finally, optimisation-based methods for motion planning such as MPC can encode the driving situation into a cost function and optimise it in a receding horizon for each time step without requiring any high-level behaviour layer~\cite{liu2017path,cardoso2017model}. The main bottleneck of this approach is its heavy computational overhead especially for increasing time horizons because all possible trajectories, even those corresponding to unrealistic manoeuvres, must be considered.

\subsection{Positioning the contributions of this paper}
The literature review reveals a need for a general-purpose motion planning method with the following properties.
\begin{itemize}
    \item A general-purpose approach that is able to solve motion planning problems for various unforeseen ADFs without needing predefined driving scenarios.
    \item It must be able to generate tactical decisions and trajectories to enable the 
    EV to safely and optimally navigate through moving/static obstacles in various road layouts.
    \item It must be computationally efficient for real-time operation using contemporary hardware.
\end{itemize}

The proposed motion planning solution for autonomous vehicles (AVs) in this paper is inspired by the theory of \textit{fluid flow} motion, which is a novel approach in the ADS domain. This technique was first used for mobile robot path planning by Keymeulen {\it {et al.}}~\cite{keymeulen1994fluid} to address the drawbacks of the potential field method. The authors of~\cite{keymeulen1994fluid} proposed a solution that generated a collision-free vector field of fluid flow from the source to the sink (the starting and end points of the manoeuvre). A similar approach was later applied in~\cite{li1998robot} for point-mass robot path planning using Poisson's (instead of Laplace's) equation to calculate the fluid flow velocity vector field. Moreover, in~\cite{you2017path}, the LBM was used to simulate the fluid flow motion at reduced computational complexity and calculate the optimal path for ships and underwater AVs~\cite{yao2018three}. The main objective of the abovementioned techniques was to find the geometric path for mobile robots. 

To the best of our knowledge, only the studies in~\cite{sulkowski2019dynamic, sulkowski2022autonomous} used the fluid flow motion to generate trajectories for AVs. In particular, the steering angle and velocity commands were derived from the distribution function of the fluid density calculated via the LBM, but only for lane-level tasks and without considering the future states of other road users, as we will do in this paper. Unlike~\cite{sulkowski2019dynamic,sulkowski2022autonomous} we consider the temporal aspects too and solve the LBM in the spatiotemporal domain, which allows to identify feasible trajectories without suffering much from performance losses, i.e., the identified trajectories are likely to be near-optimal. The fluid flow model has also been used in~\cite{sormoli2023novel} for trajectory prediction in highway scenarios, but the generated vector field over there doesn't have the temporal component.

\section{System Model and Problem Statement}
\label{sec: problem definition}
The motion planning problem considered in this study includes both tactical decision-making and trajectory generation. Since the high-level route planning does not belong to the scope of this work, it is assumed that the coordinates of the centre points of the driveable area from the start to the end of the route are fed in advance into the motion planner. Those points just define the drivable area and the calculated trajectories by the motion planner are not constrained to contain these points. Similarly, other processes implemented in separate modules and interacting with motion planning, such as object detection, velocity estimation, and trajectory prediction of surrounding vehicles (SVs)  are not developed herein either. To avoid unnecessary complexity, it is assumed that the present, $t=0$, and future states of SVs can be perfectly estimated along a time horizon of $t_p$ time steps. Specifically, the states of $n$ road objects, including static obstacles and moving road participants, are provided as input to the motion planner, i.e., $\left\{O_1\left( t \right),O_2\left( t \right), \ldots ,O_n\left( t \right)\right\}$ for $t=\{0,1,2,\ldots t_p\}$, where, $O_i(t)$ is the position of the $i$-th object after $t$ time steps in the EV's coordinate system ($x,y$). 

\begin{figure}[t!]
\centering
\includegraphics[width=1\linewidth]{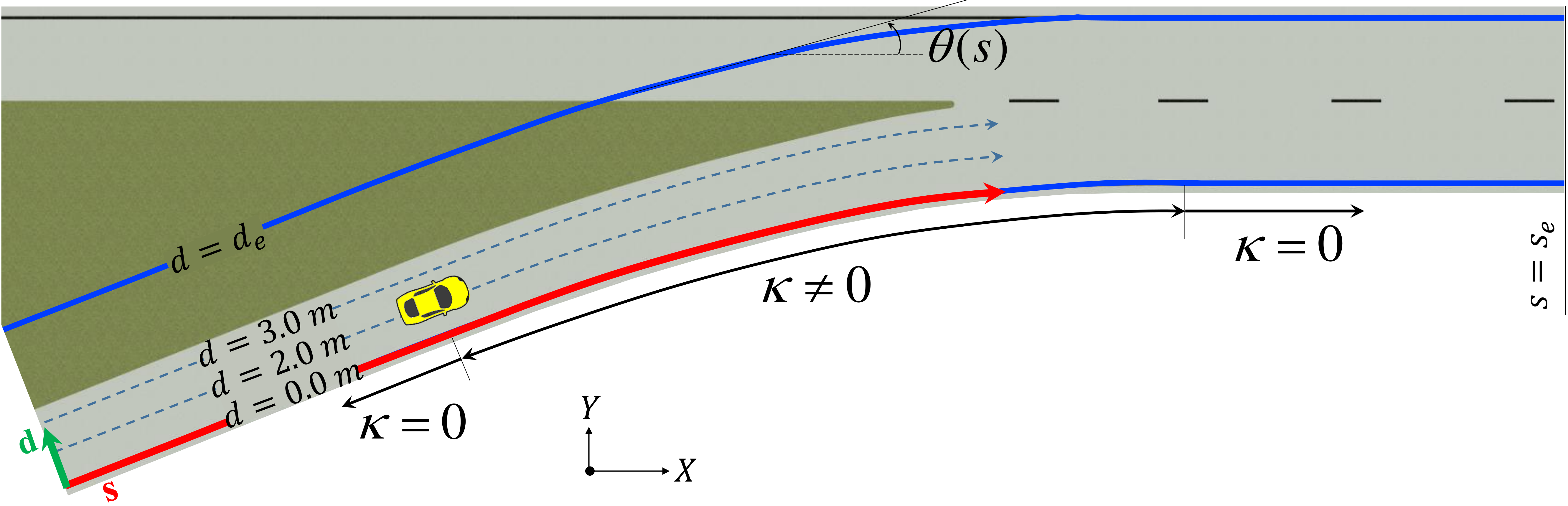}
\caption{Example illustration of Cartesian ($X,Y$) and Frenet frame ($s,d$) coordinates in a motorway merging driving scenario. One may notice the variations of the curvature $\kappa$ along the route. The parameters $s_e$ and $d_e$ denote the spatial extent of the generated trajectory in Frenet frame coordinates. The parameter $\theta(s)$ stands for the orientation of the Frenet frame with respect to the global coordinate system $(X,Y)$. }
\label{fig: Frenet frame}
\centering
\end{figure}

\begin{figure}[t!]
\centering
\includegraphics[width=.65\linewidth]{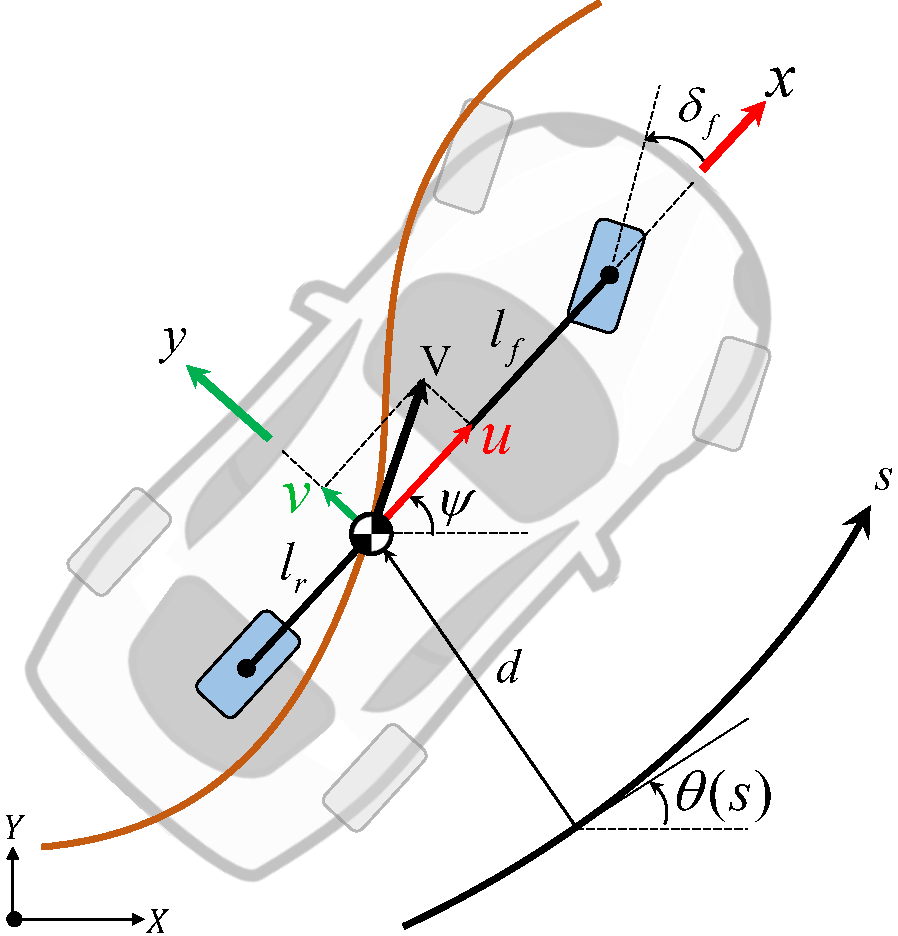}
\caption{Kinematic variables and parameters of the dynamic bicycle model used in the equations of motion. The parameter $\delta_f$ is the steering angle of the front wheels, $l_i, i\in\{f,r\}$ is the distance between the centre of gravity and the front $(i=f)$ or the rear wheels $(i=r)$, $C_i$ is the cornering stiffness constant for the wheels, and $\mu_i$ is the wheel friction coefficient. $(X,Y)$ is the global coordinate system and $(x,y)$ refers to the local coordinate system.} 
\label{fig: dynamic model}
\centering
\end{figure}

A motion planning algorithm shall calculate the EV state, ${\rm E}_t$, for $t_p$ time steps in the future, i.e., ${{\rm E}_t} \buildrel \Delta \over = (x_t,y_t,\psi_t ,\dot x_t,\dot y_t,\dot \psi_t)_{t=1}^{t_p}$, where $(x_t,y_t)$ and $\psi_t$ are the location and the yaw angle of the EV in the local coordinate frame at time step $t$. Hereafter in this section, the time index is dropped for brevity. For driving along curved roads, converting Cartesian to Frenet frame coordinates ($s,d$), see Fig.~\ref{fig: Frenet frame}, may simplify the motion planning calculations. In order to generate feasible trajectories and successfully navigate through static and moving obstacles, the nonlinear dynamics and nonholonomic constraints of the EV should be taken into account. Feasibility means that the EV should be able to follow the generated trajectory with minimum deviation and without encountering saturation in its actuators. Specifically, at each time instance $t$, the calculated trajectory should satisfy the nonlinear bicycle model~\cite{gillespie1992fundamentals} given the  present state ${\rm E}_0$:
\begin{subequations}
\label{eq: dynamic eq}
   \begin{flalign}
   \label{eq: dynaimc eq1}
  \dot u &= vr + \frac{{{F_x}}}{m}. &
 \end{flalign}  
 \begin{flalign}
 \label{eq: dynaimc eq2}
     \dot v &=  - ur + \frac{{{C_f}{\mu _f}}}{m}{\delta _f} - \frac{1}{m}\Bigg(
{C_f}{\mu _f}{\tan ^{ - 1}}\left( {\frac{{v + {l_f}r}}{u}} \right) + &\nonumber  \end{flalign} 
 \begin{flalign}
& \quad \quad \quad {C_r}{\mu _r}{\tan ^{ - 1}}\left( {\frac{{v - {l_r}r}}{u}} \right)
\ \Bigg). &
 \end{flalign} 
 \begin{flalign}
 \label{eq: dynaimc eq3}
    \dot r &= \frac{{{l_f}{C_f}{\mu _f}}}{{{I_z}}}{\delta _f} - \frac{1}{{{I_z}}}\Bigg(
{l_f}{C_f}{\mu _f}{\tan ^{ - 1}}\left( {\frac{{v + {l_f}r}}{u}} \right) - \nonumber & \\
& \quad \quad \quad {l_r}{C_r}{\mu _r}{\tan ^{ - 1}}\left( {\frac{{v - {l_r}r}}{u}} \right) \Bigg),&
\end{flalign} 
\end{subequations}
where $u$, $v$, and $r$ are the longitudinal, lateral, and yaw angular velocities, respectively ($u = \dot x$, $v = \dot y$, $r = \dot \psi$), $m$ is the mass of the vehicle, $F_x$ is the longitudinal tire force, $I_z$ is the second moment of inertia around the z-axis, and the rest of parameters are defined in the caption of Fig.~\ref{fig: dynamic model}. 

\begin{figure*}[t]
\centering
\includegraphics[width=.95\linewidth]{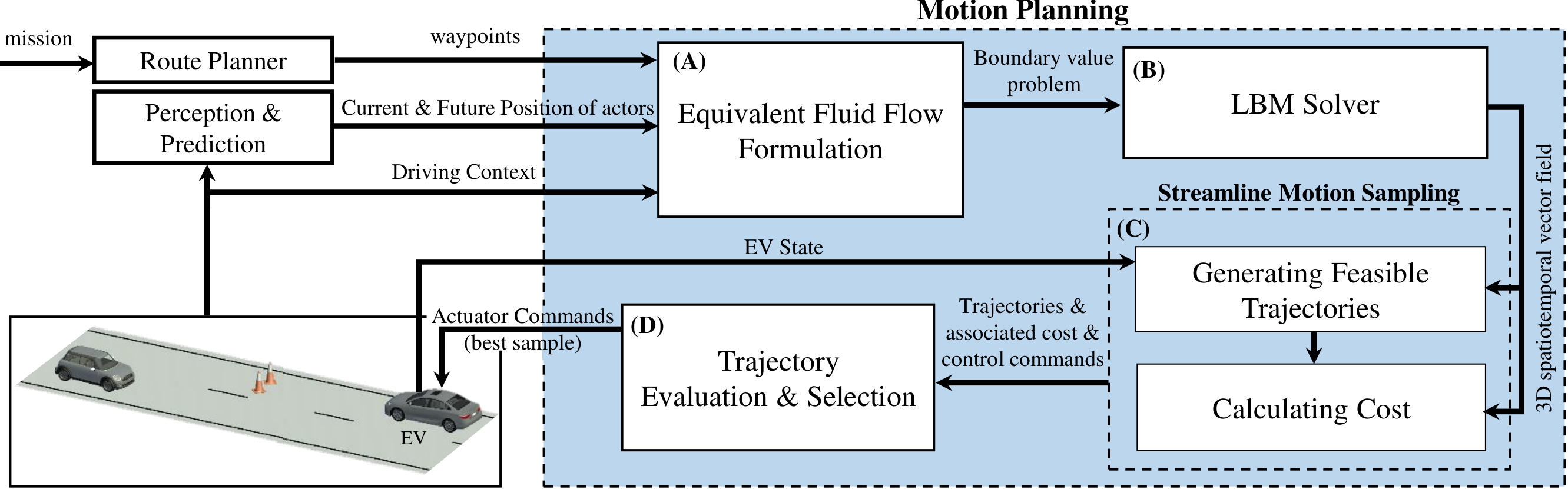}
\caption{The motion planning system diagram developed in this paper illustrating the relations between its various components. The high-level route planning along with the current and future states of surrounding vehicles (trajectory predictions) are fed into the motion planner by other modules. } 
\label{fig: mp_diagram}
\centering
\end{figure*}
\begin{figure}[t]
\centering
\includegraphics[width=1\linewidth]{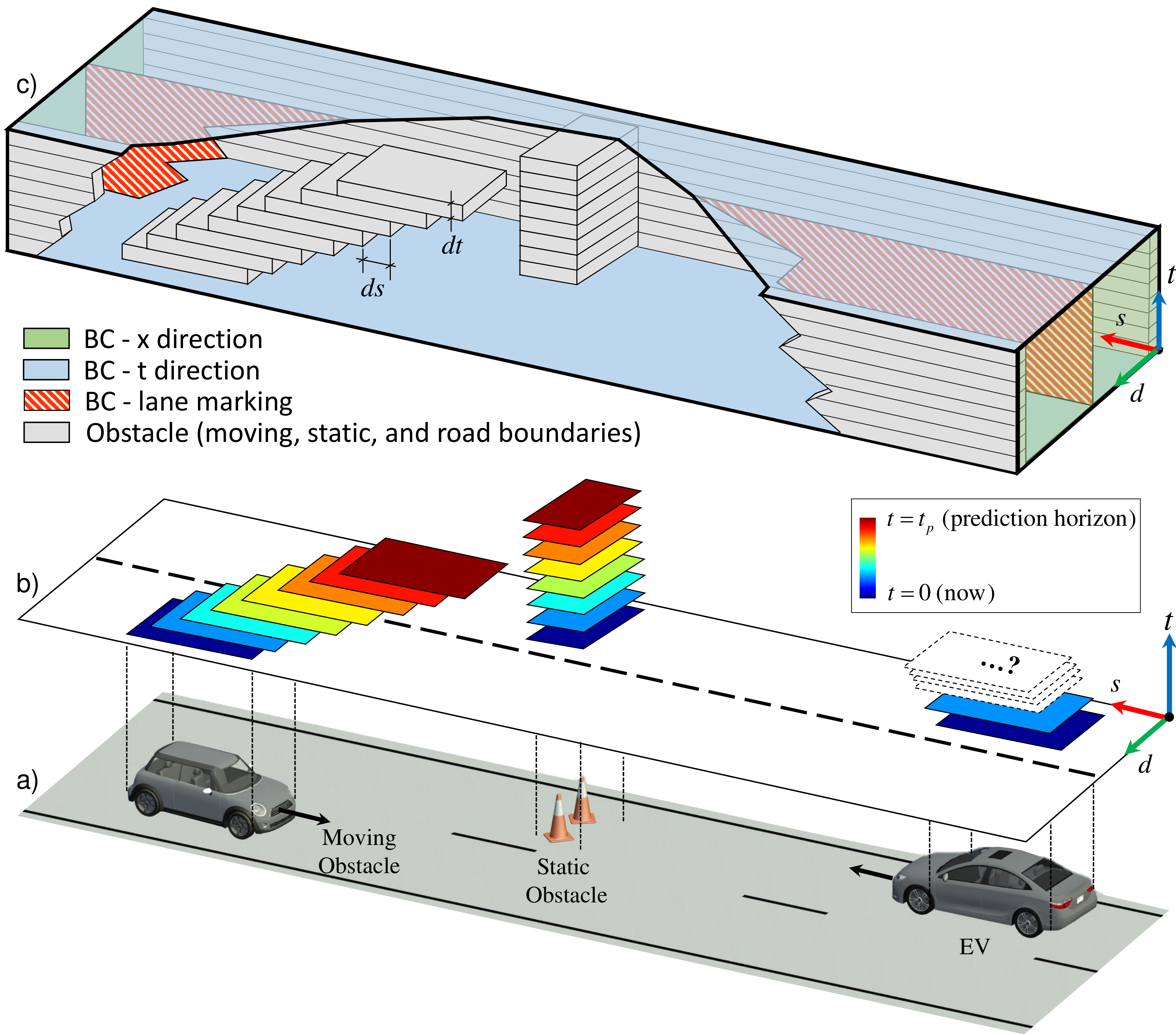}
\caption{Generating the equivalent fluid motion boundary problem (c) from the spatiotemporal representation of a 2D driving scenario (a) and (b).}
\label{fig: spatiotemporal fluid}
\centering
\end{figure}

Ideally, motion planning algorithms for AVs shouldn't be tailored to particular driving scenarios such as intersection crossing, motorway merging, etc., but instead, they should be applicable to a broad range of driving situations, including rare unknown events. That calls for general-purpose motion planning and control solutions, which cannot be obtained by utilising predefined discretised driving states and transition rules like finite state machines. For the same reason, short-term trajectory generation is not sufficient either and long-term decision-making is also required. Last but not least, the generated trajectory must be updated in real-time. Due to the fact that the update rate for contemporary sensors are the bottleneck nowadays, it is assumed that the trajectory generation processing time is constrained to $100$~ms (same as the sensor update time period~\cite{ignatious2022overview}).

\section{Proposed motion planning method}
\label{sec: proposed mp}

The fluid-inspired trajectory generation procedure developed in this paper is depicted in Fig.~\ref{fig: mp_diagram}. It consists of the following four main components:
\begin{enumerate}[label=(\Alph*)]
    \item {\it{Fluid flow formulation}}: Generates the fluid flow problem associated with the driving context. That includes the construction of the domain where the fluid flows along with the boundary conditions (BCs). 
    \item {\it{LBM solver}}: Solves the fluid flow boundary value problem using the lattice Boltzmann method (LBM).
    \item {\it{Streamline motion sampling}}: Utilises the 3D spatiotemporal vector field (STVF) obtained by the LBM to generate candidate trajectories respecting the nonlinear dynamics of Eq.~\eqref{eq: dynamic eq}. 
    \item {\it{Trajectory selection}}: Evaluates the candidate trajectories, and selects the one optimising a cost function, which factors in safety, feasibility, efficiency and comfort. 
\end{enumerate}

Each component will be explained throughout the subsequent sections. In a nutshell, the {\it{fluid flow formulation}} component (see Section~\ref{sec: equivalent fluid problem}) takes as input the high-level route planning, the current state of the environment (perception) and the prediction of the future states of other road users. It uses that to construct a boundary value problem within the equivalent fluid flow domain, which is used by the {\it{LBM solver}} (see Section~\ref{sec: LBM}) to produce the associated 3D STVF. The {\it{streamline motion sampling}} component (see Section~\ref{sec: Streamline sampling} and Algorithm~\ref{alg: sampling}) uses the STVF and the current state of the EV to sample $N$ feasible trajectories using Eq.~\eqref{eq: dynaimc eq1}-~\eqref{eq: dynaimc eq3} while also considering the saturation of the EV actuators. The associated control signals (throttle/brake and steering angle) for each trajectory are also recorded. The candidate trajectories are compared within the {\it{trajectory selection}} component (see Section~\ref{sec: costFunction}). The first state of the best trajectory becomes the new state for the EV and the above process repeats.  
\begin{figure*}[t]
\centering
\includegraphics[width=1\linewidth]{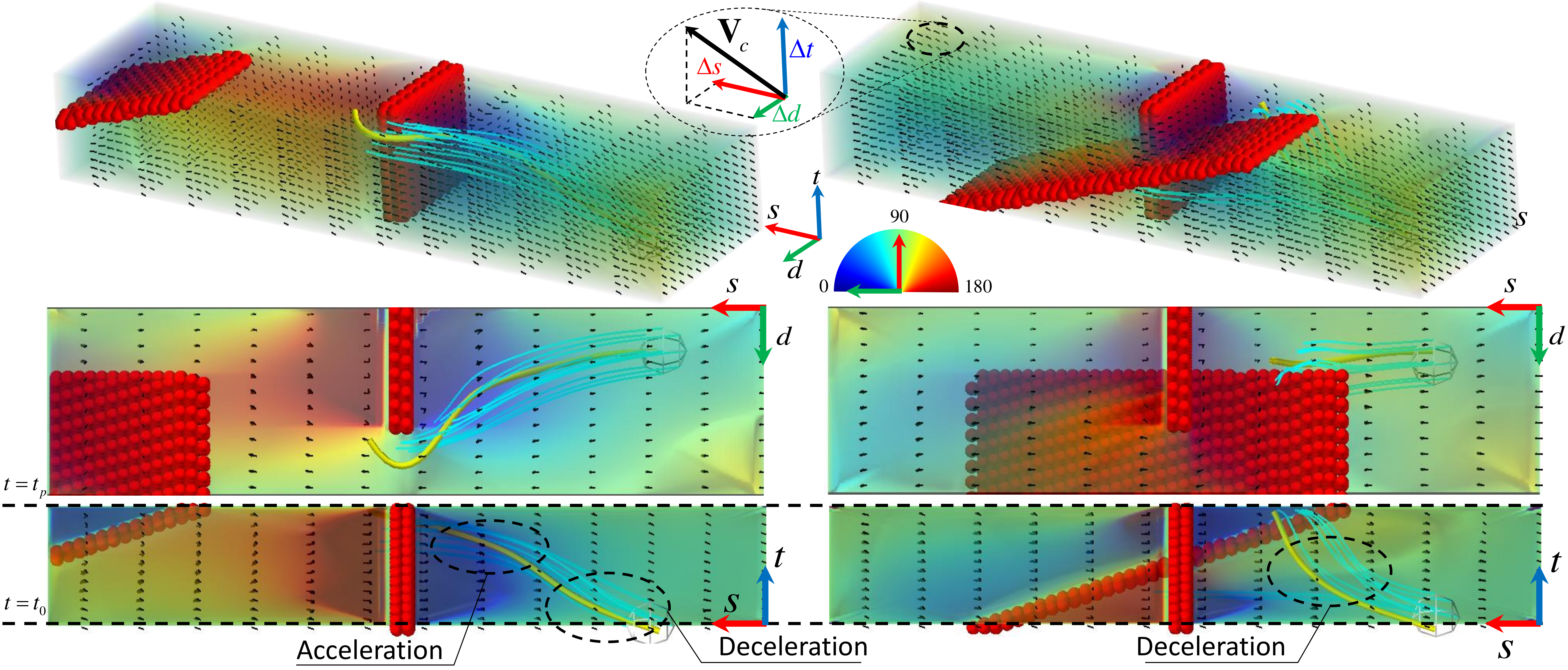}
\vspace{-4mm}
\caption{Example illustrations of a STVF and  sample streamline trajectories generated for two different cases of the scenario presented in Fig.~\ref{fig: spatiotemporal fluid}. The EV overtakes the static obstacle  considering another vehicle moving in the opposite direction at far (left) and close (right) ranges. The STVF (illustrated by the black arrows and the interpolated cyan streamlines) guides the EV (yellow curve) to accelerate and overtake the static obstacle before/after the moving vehicle arrives (left/right) leading to an initial deceleration in the latter case (right). In this illustration, the trajectory of the EV (yellow curve) does not exactly follow one of the fluid streamlines (cyan curves) because of saturation in its control signals. The colourmap corresponds to the lateral orientation of the STVF in the $(s,d)$ plane which is an indication of a left/right turn.}
\vspace{-5mm}
\label{fig: stvf}
\centering
\end{figure*}

\subsection{Fluid flow formulation}
\label{sec: equivalent fluid problem}
To convert the driving context into the equivalent fluid flow boundary value problem, we first need to define the fluid motion domain denoted by $\Omega$. This has the form of a rectangular cuboid (see Fig.~\ref{fig: spatiotemporal fluid}c for an illustration) that has two dimensions ($s$ and $d$) representing the drivable area in Frenet coordinates and the third dimension ($t$) representing time. The length, width, and height of $\Omega$  determine the spatial extent of the road and the prediction horizon, respectively, i.e., $s\in (0,s_e)$, $d\in (0,d_e)$, and  $t\in(0,t_p)$.

In practice, the domain $\Omega$ is discretised using cuboid cells, and each cell would be associated with a 3D vector (spatiotemporal increment) giving rise to a 3D STVF across the domain. The origin of each vector is placed at the centre of the cell, and for simplicity, the vector magnitude is normalised to unity. If the spatiotemporal displacement vector is denoted by $(\Delta s,\Delta d,\Delta t)$, the instantaneous speed in Frenet coordinates can be calculated as $\dot{s} = \Delta s/{\Delta t}$ and $\dot{d} = {\Delta d}/{\Delta t}$. In order to calculate the STVF, we first need to determine the BCs for the six sides of $\Omega$, i.e., the 3D  vectors associated with the cuboid cells residing at the sides of the fluid flow domain. This is discussed next.  

 Intuitively, for the bottom side representing the time step $t=0$,  the current velocity of the EV is used, while for the top side $t=t_p$, the velocity is set equal to the \textit{nominal velocity} of the related part of the road, unless the corresponding spatiotemporal locations are already occupied by some obstacle (moving or static). For the top side, it is natural to assume that the EV should attain the nominal speed only at the $s$-direction (i.e., along the direction of the road), while its speed at the $d$-direction is zero, i.e., ${\Delta d}=0$. For the remaining sides $s=0$ and $s=s_e$ (excluding the bottom side cells $t=0$) the nominal speed (with vector component $\Delta d=0$) is used as well, while for the sides $d=0$ and $d=d_e$ the STVF is set to zero so that the EV stays within the road boundaries. 

Boundary conditions may also arise within the fluid flow domain. For instance, the current and future positions of moving and static obstacles translate into the \textit{occupied space} within $\Omega$, see the grey-coloured regions in Fig.~\ref{fig: spatiotemporal fluid}c, where fluid particles are not allowed to flow. Hence, the cuboid cells within the occupied space are assigned zero vectors. Finally, \textit{lane marking} is another semantic information that needs to be converted in the equivalent fluid motion problem. Since the EV should be able to cross lane markings (unlike road boundaries), a porous medium is used to model them. To create it, only some of the cuboid cells residing along the  porous are assigned a zero vector to impede (to some extent) the fluid motion. Apparently, the percentage of these cells across the porous reflects the resistance ($R_p$) of the porous, which describes how difficult it is for fluid particles to move from one side of the porous to the other.

After defining the fluid flow domain $\Omega$, its BCs, the porous resistance and the occupied portions, the fluid dynamic equations of motion (also known as Navier-Stokes equations, see Appendix~\ref{app: navier-stokes} for more details) can be numerically solved to calculate the STVF within the domain $\Omega$ and describe the fluid flow motion~\cite{glowinski2003numerical}. 


\subsection{Lattice Boltzmann method (LBM)}
\label{sec: LBM}
Due to the lack of analytic formulas for obtaining the solution to the Navier-Stokes equations, we had to carefully consider the pros and cons of several numerical routines available in the literature~\cite{bailey2009accelerating, januszewski2014sailfish, behr1994finite}. Among them, the LBM has been finally selected, as it does not only handle BCs with sophisticated geometries but also allows parallel computations using graphics processing units (GPUs). Both features make the LBM  suitable for a motion planning application in which computational efficiency and ingestion of complex driving contexts/domains are key requirements. 

According to the LBM, the simulated fluid volume $\Omega$ is first discretised into uniformly spaced grids on a lattice (cuboid cells for the 3D domain in our case study). After assigning the BCs as described in the previous section, the LBM indirectly solves the associated differential equations and calculates the motion vectors via two main steps, propagation (streaming) and collision (relaxation) of fluid density in the lattice. The LBM applies an iterative process where the microscopic fluid density propagates along the lattice and the lattice densities are updated through the collision process (details are discussed in Appendix~\ref{app: lbm}). In the end, the  STVF, denoted by $\mathcal{V}$, is obtained. It consists of normalised 3D vectors with unit magnitude, $V_c=(\Delta s_c, \Delta d_c,\Delta t_c): \lVert V_c \rVert_2=2$, with the subscript indicating the $c$-th cuboid cell in the domain $\Omega$, see Fig.~\ref{fig: stvf} (top-right) for an example  illustration. 

 Since the core process of the motion planning method is the fluid motion simulation, the computational complexity of the LBM can be used to infer the implementation efficiency of the motion planning algorithm.  That can be measured in terms of million lattice updates per second (MLUPS). 
 \textit{Sailfish} is a well-developed open-source toolbox that implements the LBM  with good enough flexibility for defining a problem and incredible computational performance on GPU (up to 4000 mlups on GeForce RTX 3080 Ti)~\cite{januszewski2014sailfish}. Therefore, if it takes 100 lattice updates to calculate the STVF, the computation takes only $14$~ms for a lattice with dimensions 128$\times$64$\times$64, which is much smaller than the sensor update rate (10~Hz).

Before completing this section, we determine the dimensions of the domain $\Omega$ and cuboid cells for our application. Recall that the domain of the fluid flow simulation represents the Frenet frame of the driving route. The length of this volume corresponds to the road length that is set equal to $s_e=256$~m in our case. It starts $30$~m behind and extends up to $226$~m in front of the current position of the EV. The width of the domain is equal to the road width along the driving route, which may vary from the width of a single to that of multiple lanes. Two-lane roads with a total width equal to $d_e=6.4$~m is assumed here. Finally, the third dimension (height) represents time, which starts at $t=0$ (current time step) and extends to the prediction horizon $t_p=6.4$~s. As illustrated in Fig.~\ref{fig: spatiotemporal fluid}c, the origin of $\Omega$ is at the corner of the simulation domain. Moreover, in this study, the number of cuboid cells within $\Omega$ are 128$\times$64$\times$64 in length, width, and height, respectively or $c=1,2,\ldots, 2^{19}$ in total. This translates to a cell size in longitudinal, lateral, and time dimensions equal to $2.0$~m, $0.1$~m, and $100$~ms, respectively.

\subsection{Streamline motion sampling}
\label{sec: Streamline sampling}

Considering the current state ${{\rm E}_0} \buildrel \Delta \over = (x_0,y_0,\psi_0 ,u_0, v_0,\dot \psi_0)$, and given the STVF obtained by the LBM, in this section, we show how to sample feasible trajectories for the EV. Intuitively, the EV shall follow the fluid flow minimising the energy consumption while complying with the equations of motion given in Eq.~\eqref{eq: dynamic eq}. However, the EV motion may not follow exactly the STVF due to potential saturation of control inputs or slip angles out of bounds, see the yellow curves in Fig.~\ref{fig: stvf}. Because of that, we generate more trajectories around the fluid-based particle trajectory to increase the probability of finding the optimum trajectory. The trajectory generation process is described in the form of pseudocode under Algorithm~\ref{alg: sampling} and explained below. 

The inputs of Algorithm~\ref{alg: sampling} in the current time step are the state of the EV and the 3D STVF. Furthermore, the number of candidate trajectories $N$ is set. Since the dynamic equations of motion are in the EV local frame $(x,y)$, but the STVF is based on the road's Frenet frame $(s,d)$, coordinate transforms are required in the trajectory sampling procedure (\textbf{lines~10 and 14}). Assuming a time granularity ${\rm d} t$, the location of the EV in the next time step $(t+1)$ is estimated based on its current location and velocity  (\textbf{line~11}). The velocity of the EV in the new location is obtained by trilinear interpolation of the vectors of the STVF belonging to the nearest and all neighbouring cuboid cells  (\textbf{line~12} with $g(\cdot)$ being the interpolating function). Over there the scalars $\{\gamma_i,\eta_i\}$ are appropriately defined for $i\neq 1$ so that the generated trajectories are diverse. Given two consecutive velocities, the acceleration in the Frenet frame is obtained considering Coriolis and centrifugal components (\textbf{line~13} with $\kappa$ being the curvature of the reference profile) and transferred to the EV frame (\textbf{line~14}). Note that numerical derivation often amplifies the noise and in practice, a digital lowpass filter, such as the Savitzky–Golay~\cite{schafer2011savitzky}, would be required to smooth the resulting accelerations. 
By substituting the accelerations in Eq.~\eqref{eq: dynamic eq}, the associated control inputs $(F_x,\delta_f)$ are obtained (\textbf{line~15}) using inverse dynamics, denoted by $\mathcal{F}^{-1}(\cdot)$, and subsequently checked for saturation (\textbf{line~16}).  
Finally, the rate of change in the state of the vehicle using the control signals $(F_x, \delta_f)$ is obtained using forward dynamics, denoted by $\mathcal{F}(\cdot)$, (\textbf{line~17}). After that, the new state of the vehicle is appended to the trajectory array (\textbf{line~18}), and the control inputs are appended in the array of control inputs (\textbf{line~19}).
For completeness, the forward and inverse dynamics are explained in Appendix~\ref{app: inverse dynamic}. 

\newcommand{\xCommentSty}[1]{\scriptsize\ttfamily\textcolor{blue}{#1}}
\SetCommentSty{xCommentSty}

\setlength{\algomargin}{0pt}

\RestyleAlgo{ruled}
\SetKwComment{Comment}{/* }{ */}
\SetKw{KwBy}{by}
\SetKwInOut{Input}{Input}
\SetKwInOut{Output}{Output}
\IncMargin{1.3em}

\begin{algorithm}[t!]
\caption{Vector field guided trajectory sampling}\label{alg: sampling}

\Input{%
${{\rm{E}}_0} = {({x_0,y_0,\psi_0,u_0,v_0,\dot \psi_0})}$ -- Initial state\\
$\mathcal{V}$ -- STVF within $\Omega$\\
$V_c = (\Delta s_c,\Delta d_c,\Delta t_c) : \lVert V_c\rVert_2=1, \forall c \in \Omega$ \\ 
$N$ -- number of sampled trajectories \\ 
${\rm d}t = 100$~ms -- time granularity\\ 
} 

\Output{$\mathcal{T}_i, i=1,2,\ldots, N$ -- Trajectory sample\\
$\mathcal{I}_i , i=1,2,\ldots, N$ -- Control inputs 
}

\BlankLine

\BlankLine
\For{$i\gets 1$ \KwTo $N$ \KwBy $1$}{

\eIf {$i=1$} {
$\{\gamma_1,\eta_1\} = \{1,1\}$
}{$\{\gamma_i,\eta_i\}$ -- Initialise}

 $\mathcal{T}_i \gets$ empty array \\     
 $\mathcal{I}_i \gets$ empty array  \\
 \For{$t\gets 0$ \KwTo $t_{p}$ \KwBy $1$}{
    $\left\{ \begin{array}{l}
\beta_t  = \theta_t  - \psi_t \\ 
\dot s_t \gets u_t\cos \beta_t  - v_t\sin \beta_t  \\
\dot d_t \gets u_t\sin \beta_t  + v_t\cos \beta_t 
\end{array} \right.$\  \\ 
${s_{t+1}} \gets s_t + \dot s_t {\rm d} t;\quad {d_{t+1}} \gets d_t + \dot d_t {\rm d} t$\

\BlankLine
$\left\{ \begin{array}{l}
{\dot s}_{t+1} \gets {\gamma_i} g({s_{t+1}},\mathcal{V})\\
{\dot d}_{t+1} \gets {\eta_i} g(d_{t+1},\mathcal{V})
\end{array} \right.$\

\BlankLine
$\left\{ {\begin{array}{*{20}{l}}
{{{\ddot s}_t} = ({{\dot s}_{t+1}} - {{\dot s}_t})/{{\rm d}t} + 2\kappa {{\dot s}_t}{{\dot d}_t}}\\
{{{\ddot d}_t} = ({{\dot d}_{t+1}} - {{\dot d}_t})/{{\rm d} t} + 2\kappa \dot s_t^2}
\end{array}} \right.$

\BlankLine
$\left\{ {\begin{array}{*{20}{l}}
{{{\dot u}_t} \leftarrow {{\ddot s}_t}\cos {\beta _t} + {{\ddot d}_t}\sin {\beta _t}}\\
{{{\dot v}_t} \leftarrow {{\ddot d}_t}\cos {\beta _t} - {{\ddot s}_t}\sin {\beta _t}}
\end{array}} \right.$\   
 
 \BlankLine
 $\left( {{F_x},{\delta_f}} \right) \leftarrow \mathcal{F}^{-1}\left({\dot u}_t,{\dot v}_t\right)$ 
 
 \BlankLine
 $\left( {{F_x},{\delta _f}} \right) \gets$  ${\text{Saturate}} \left(F_x,\delta_f \right)$ \
 
 \BlankLine
 $\left({\dot u}_t,{\dot v}_t\right) \leftarrow \mathcal{F} \left( {{F_x},{\delta _f}} \right)$
 
 \BlankLine
$\mathcal{T}_i \gets$ Append ${\rm E}_t$ to $\mathcal{T}_i$\

$\mathcal{I}_i \gets$ Append $\left( {{F_x},{\delta _f}} \right)$ to $\mathcal{I}_i$\ \\ 
\BlankLine
   } 
    }
\end{algorithm}

\subsection{Trajectory selection}
\label{sec: costFunction}
With the calculation of $N$ trajectories at hand, $\mathcal{T}_i, i=1,2,\ldots, N$, we need to assess the quality of each trajectory and select the optimal in terms of suitable key performance indicators (KPIs). One way to do the comparison is to assign a scalar value $J$ to each candidate trajectory that is a measure of its quality (the lower, the better). In our case, the quality is determined based on the cumulative summation of safety, comfort, and control effort KPIs.
\begin{equation}
\label{eq: costFunction}
J = \sum\limits_{t=0}^{{t_p}} \left( {{J_1}(t) + {J_2}(t) + {J_3}(t) + {J_4}(t)} \right).
\end{equation}

In AV applications the instantaneous safety, $J_1(t)$, is often evaluated based on the time headway, the post-encroachment time, and/or the time-to-collision (TTC) between the EV and moving/static obstacles~\cite{Hazim2023}. Motivated by the formulation of motion planning as a fluid flow problem, we hereby suggest using the shear stress to describe safety. Shear stress is proportional to the velocity difference between sliding layers of the fluid particle~\cite{glowinski2003numerical}, and according to the no-slip condition in fluid motion simulation, the shear stress becomes maximum near the obstacles where a boundary layer is formed. Therefore, lower shear stress necessitates larger spatiotemporal distances between the EV and obstacles in the driving scene, which is associated with safer driving. 
\begin{align}
{J_1}(t) = c_1 h({s_t},{d_t},\mathcal{V})^2,
\end{align}
where $c_1>0$ is a weighting coefficient and $h$ is the shear stress function.

The instantaneous passenger comfort $J_2(t)$ is commonly assessed using the magnitude of the instantaneous accelerations or the jerk in the lateral and longitudinal plane of motion (lower accelerations are associated with improved ride quality)~\cite{Hazim2023}. For motion planning, the square of the lateral and longitudinal accelerations are preferred over their absolute values in order to generate a cost function that is differentiable. Hence, 
\begin{equation}
    {J_2}(t) = \begin{bmatrix}
{{{\dot u}_t}}&{{{\dot v}_t}}
\end{bmatrix} \begin{bmatrix} {{c_{21}}}&0\\
0&{{c_{22}}}
\end{bmatrix} 
\begin{bmatrix}
{{{\dot u}_t}}&{{\dot v}_t} \end{bmatrix}^T,
\end{equation}
where $c_{21}, c_{22}$ are positive scalars.

Finally, the efficiency can be measured by the control effort that is quantified by the magnitude of the longitudinal force and steering angle scaled by some constants.
\begin{align}
{J_3}(t) = \begin{bmatrix} F_x(t) & \delta_f(t)
\end{bmatrix} \begin{bmatrix} {{c_{31}}}&0\\ 0&{{c_{32}}}
\end{bmatrix} \begin{bmatrix} {{F_x(t)}}&{{\delta_f(t)}}
\end{bmatrix}^T,
\end{align}
and similarly for their rate of change 
\begin{align}
{J_4}(t) = \begin{bmatrix} {{\dot F_x}}&{{\dot \delta_f}}
\end{bmatrix} \begin{bmatrix} {{c_{41}}}&0\\
0&{{c_{42}}} \end{bmatrix} 
\begin{bmatrix}  {{\dot F_x}}&{{\dot \delta_f}} \end{bmatrix}^T.
\end{align}

\section{Performance Evaluation}
\label{sec: evaluation}
The performance of the proposed motion planning method is evaluated in this section using simulations. Various driving scenarios (ADFs) are implemented, including different road layouts and road participant behaviours, which are described in Section~\ref{sec:IV_A}. A receding horizon optimization-based method, also known as model predictive control (MPC), is utilised as the baseline for comparison. MPC is a natural choice to represent the performance of state-of-the-art motion planners because it identifies a local optimal solution given the initial state of the EV, the cost function and the reference manoeuvre.   It is worth clarifying that the developed motion planner doesn't require a separate decision-making mechanism. This is in stark contrast with MPC, where we generate the optimal trajectory for each possible manoeuvre/decision and that associated with the best performance is finally selected as the baseline output~\cite{rasekhipour2016potential}. Since the best decision is not known in advance, the reported performance of MPC essentially corresponds to an upper bound. Furthermore, MPC is computationally expensive and thus, it's not a realistic choice for real-time operations. For the implementation details of MPC see~\cite{rasekhipour2016potential}, which is the reference for designing MPCs in our previous works~\cite{mozaffari2023multimodal, mozaffari2023trajectory}. In Section~\ref{section: metrics}, the trajectories generated by the two motion planners are post-processed and compared in terms of key performance indicators (KPIs) capturing safety, comfort, efficiency, computational cost, and feasibility of throttle/brake and steering control signals. Presentation and discussion of the simulation results follow in Section~\ref{sec: sim results}. 


\begin{table*}
\SetTblrInner{row{3,5,7,9,11,13,15}={gray9}}
\SetTblrInner{cell{3,7,9}{1,1,1}={white}}
\centering
\caption{Simulation parameters of actors (EV and OVs) for three different initial conditions per evaluation scenario. \textbf{p1}: initial position $({x_0},{y_0})$ of the actor, \textbf{p2}:  initial speed and heading $({V_0},{\psi}_0)$, \textbf{p3}: acceleration in Frenet frame: $({\ddot d},{\ddot s})$, all in SI units. The initial position of the EV is the origin.}
\vspace{-.2cm}
\label{table: scenarios}
\begin{tblr}{
  cells = {c},
  cell{1}{3} = {c=3}{},
  cell{1}{6} = {c=3}{},
  cell{1}{9} = {c=3}{},
  cell{1}{12} = {c=3}{},
  cell{4}{1} = {r=3}{},
  cell{7}{1} = {r=3}{},
  cell{10}{1} = {r=3}{},
  hline{1,3,4,7,10,13} = {-}{},
  hline{2} = {3-14}{},
  column{3-8}={c}{.9cm},
  column{1}={c}{.6cm},
}
             &             & \textbf{Scenario (a)} &             &             & \textbf{Scenario (b)} &             &             & \textbf{Scenario (c)} &             &             & \textbf{Scenario (d)} &             &             \\
             &             & \textbf{a1}           & \textbf{a2} & \textbf{a3} & \textbf{b1}           & \textbf{b2} & \textbf{b3} & \textbf{c1}           & \textbf{c2} & \textbf{c3} & \textbf{d1}           & \textbf{d2} & \textbf{d3} \\
\textbf{EV}  & \textbf{p2} & (15,0)                & (15,0)      & (15,0)      & (25,30)               & (25,30)     & (25,30)     & (5,90)                & (5,90)      & (5,90)      & (30,90)               & (30,90)     & (30,90)     \\
\textbf{OV1} & \textbf{p1} & (40,0)                & (40,0)      & (40,0)      & (-70,5)               & (-70,5)     & (5,5)       & (-30,10)              & (-50,10)    & (25,10)     & (-90,100)             & (-90,100)   & (-90,100)   \\
             & \textbf{p2} & (0,0)                 & (0,0)       & (15,0)      & (30,0)                & (30,0)      & (35,0)      & (15,0)                & (10,0)      & (15,0)      & (30,0)                & (20,0)      & (35,0)      \\
             & \textbf{p3} & (0,0)                 & (0,0)       & (-8,0)      & (0,0)                 & (2.5,0)     & (0,0)       & (0,0)                 & (-1,0)      & (1,0)       & (0,0)                 & (-2,0)      & (2,0)       \\
\textbf{OV2} & \textbf{p1} & (60,4)                & (100,4)     & (100,4)     & (50,5)                & (50,5)      & (-35,5)     & (35,14)               & (20,14)     & (50,10)     & (90,14)               & (90,14)     & (90,14)     \\
             & \textbf{p2} & (-10,0)               & (-10,0)     & (-10,0)     & (30,0)                & (30,0)      & (35,0)      & (15,180)              & (10,180)    & (10,180)    & (30,180)              & (35,180)    & (25,180)    \\
             & \textbf{p3} & (0,0)                 & (0,0)       & (0,0)       & (0,0)                 & (-2.5,0)    & (0,0)       & (-1,0)                & (1,0)       & (-1,0)      & (0,0)                 & (2,0)       & (-2,0)      \\
\textbf{OV3} & \textbf{p1} & …                     & …           & …           & (80,5)                & (80,5)      & (70,5)      & …                     & …           & …           & …                     & …           & …           \\
             & \textbf{p2} & …                     & …           & …           & (30,0)                & (30,0)      & (30,0)      & …                     & …           & …           & …                     & …           & …           \\
             & \textbf{p3} & …                     & …           & …           & (0,0)                 & (0,0)       & (0,0)       & …                     & …           & …           & …                     & …           & …                   
\end{tblr}
\end{table*}

\begin{figure*}[t]
\centering
\includegraphics[width=1\linewidth]{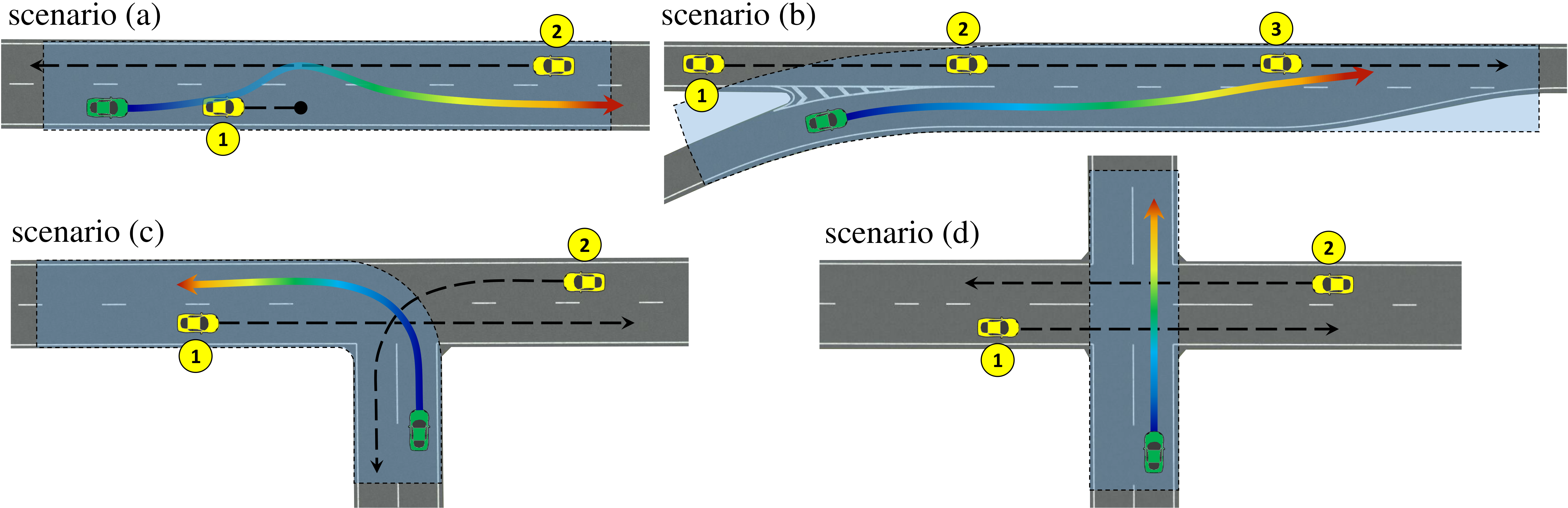}
\caption{Evaluation scenarios. Scenario (a): Straight bidirectional road on which the EV have to overtake OV1 while handling oncoming traffic. Scenario (b): Highway merging into the appropriate gap. Scenarios (c) and (d): Intersection with two effective OVs approaching the junction with different behaviours. The flow simulation domain $(s,d)$ coordinates is shown by a transparent light blue area (obtained from the high-level route planner).}
\label{fig: scenarios}
\centering
\end{figure*}

\subsection{Selection of driving scenarios}
\label{sec:IV_A}
We have designed four driving scenarios to evaluate and compare the performances of motion planning algorithms, see  Fig.~\ref{fig: scenarios}. Three use cases per scenario with different initial conditions are simulated, see Table~\ref{table: scenarios}. Since  MPC requires the output of the high-level tactical decision-making process to generate trajectories, a finite number of decisions are associated with each scenario. These are presented below. 

\textit{Scenario (a)}: The drivable area is a straight bidirectional road with two lanes and two obstacle vehicles (OVs) are effective. OV1 moves in front of the EV in the same lane with the same initial speed but decelerates to a full stop. In the other lane with opposite traffic, OV2 moves with constant speed but its initial position is different per test case. The available decisions for the EV are: (i)  Overtake OV1 before OV2 arrives at the overtaking position, and (ii) wait for OV2 to pass and then overtake OV1. 

\textit{Scenario (b)}: This is a highway merging from a slip road to the main carriageway in which three OVs move with different speeds, accelerations, and gaps in between them. For this scenario, there are three available decisions since the EV can merge into gaps after each one of the OVs.  

\textit{Scenario (c)}: The road layout is a T-junction. OV1 and OV2 approach the junction from the left and right of the horizontal bidirectional road, respectively. OV1's trajectory is straight, while OV2 takes a turn and joins the vertical road on the opposite lane of the EV. There are three available decisions for the EV: (i) Wait for both OVs to clear the junction. (ii)/(iii) OV1/OV2 clears the junction while OV2/OV1 approaches at a farther distance so that the EV can clear the junction before OV2/OV1.   

\textit{Scenario (d)}: This is similar to Scenario~(c) except that the drivable area is an intersection with four arms. Both bidirectional roads contain two lanes. The EV moves on the vertical road while the OVs move on the horizontal road from right and left with constant speeds but different initial positions. There are two available decisions for the EV: (i) Wait for the OVs to clear the intersection, or (ii) keep moving before the OVs reach the intersection. 

\begin{figure*}[t]
\centering
\includegraphics[width=.95\linewidth]{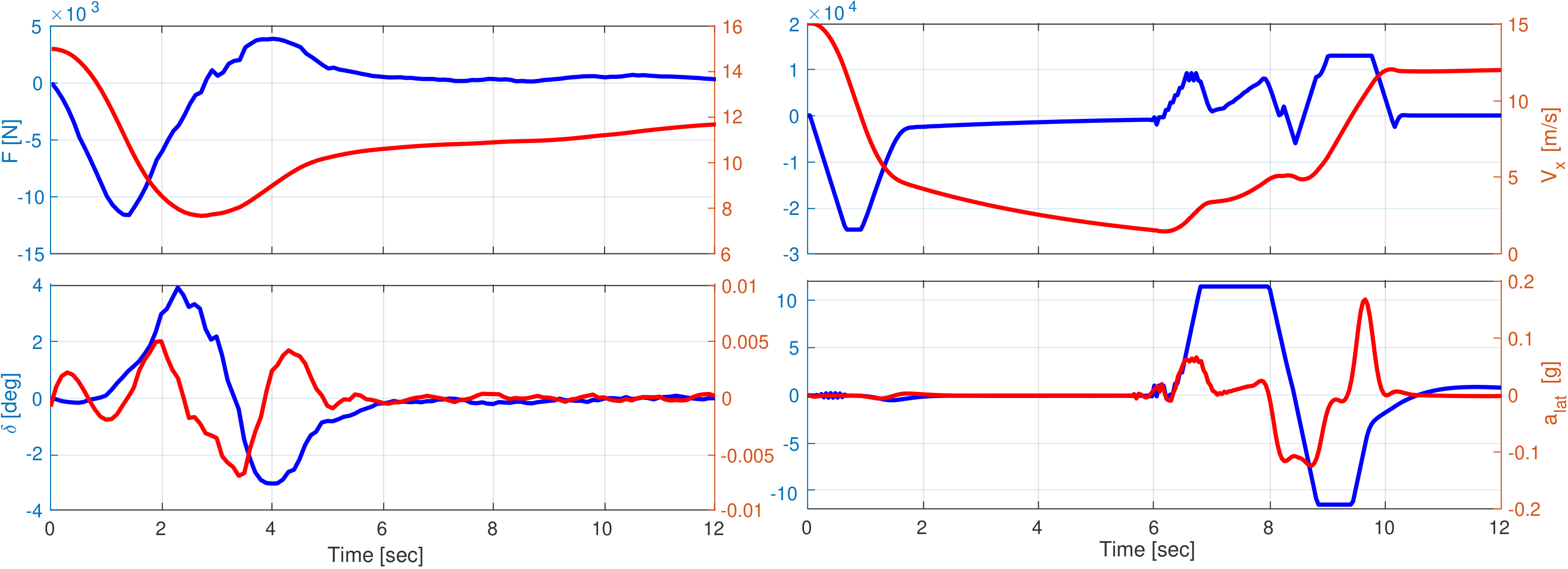}
\caption{Simulation result for qualitative comparison between the proposed method (left) and MPC-based motion planner (right) for scenario \textbf{a1}. The left y-axes show control input of throttle/brake (top) and steering angle (bottom), while the right y-axes show speed (top) and lateral acceleration (bottom).}
\label{fig: result_compare}
\centering
\end{figure*}

\subsection{Selection of performance metrics}
\label{section: metrics}
To compare motion planners, the generated trajectories for each scenario are post-processed and evaluated in terms of safety, comfort, and feasibility. For a comparative assessment, the KPIs must be unified, e.g., the shear stress used to evaluate safety before selecting the optimal trajectory in fluid-inspired motion planning (see Section~\ref{sec: costFunction}) is not applicable to MPC. The selected KPIs during post-processing are expanded below.

\subsubsection{Safety}
The time-to-collision (TTC) is a popular measure of safety in motion planning, which is calculated based on the radial relative velocities and distances between the EV and all surrounding moving objects throughout the EV's trajectory~\cite{mori2023shard}. Since the TTC can take high values and/or some of the OVs might not be on a collision course with the EV, i.e., ${\rm TTC}\to\infty$, its inverse is better suited to evaluate safety (the lower the better): 
\begin{equation}
\label{eq: safety}
    {K_{s}} = \frac{1}{{{M}\cdot{T}}}\sum\limits_{j = 1}^{{M}} {\sum\limits_{\tau = 1}^T \frac{1}{{\rm TTC}_{j,\tau}}}, 
\end{equation}
where $M$ and $T$ are the numbers of OVs and time steps in the generated trajectory, respectively, and the subscripts show that the TTC is averaged across all OVs and time steps. It shall be noted that in Eq.~\eqref{eq: safety} the time index $\tau$ has been used to highlight that it is the actual trajectory evaluated here rather than each candidate trajectory within the prediction horizon, which was indicated by another subscript $t=0,1,\ldots, t_p$ in Section~\ref{sec: costFunction}. 

\subsubsection{Comfort} 
The weighted sum of the absolute instantaneous lateral and longitudinal acceleration averaged across the trajectory  (the lower the better) is often used to assess ride quality. Hence,   
\begin{equation}
\label{eq: comfort}
    K_c = \frac{1}{T}\sum\limits_{\tau = 1}^T \left( c_{lat} |{\dot u}_\tau|  + c_{lon} |{\dot v}_\tau| \right), 
\end{equation}
where $c_{lon}<c_{lat}$ is a reasonable choice since the lateral jerk generates more discomfort to the passengers.

\subsubsection{Feasibility}
To evaluate efficiency, the saturation of the EV's control signals of throttle/brake $F_x(\tau)$ and steering angle $\delta_f(\tau)$ can be monitored. Specifically, the fraction of time where the control inputs are not saturated indicates the feasibility of a trajectory (the higher the better): 
\begin{equation}
\label{eq: feasibility}
{K_{f}} = \frac{1}{2T}\sum\limits_{\tau = 1}^T \left( \mathbb{1}\{ F_x(\tau) < F_{max}\} + \mathbb{1}\{ |\delta_f(\tau)| < \delta_{max}\} \right), 
\end{equation}
where $\mathbb{1}\{\cdot\}$ is the (binary) indicator function and the positive parameters $F_{max}$ and $\delta_{max}$ are the maximum feasible throttle force and steering angle, respectively.

\begin{table}
\centering
\caption{Quantitative comparison between MPC and proposed motion planning using metrics explained in Section \ref{section: metrics}. Values are averaged among all simulated scenarios.}
\begin{tblr}{
  column{even} = {c},
  column{3,5} = {c},
  row{1} = {m},
  hlines,
  column{2-6}={c}{1cm},
}
Method   & $\max ({K_c})$ (g) & $\max ({K_s})$ (1/s) & ${K_f}$ (\%) & $\left| F \right|$ (KN) & Time (s) \\
Proposed      & \textbf{\underline{0.007}}      & 0.43           &  \textbf{\underline{86}}     &  \textbf{\underline{4.15}}                         &  \textbf{\underline{0.083}}      \\
MPC & 0.011      &  \textbf{\underline{0.39}}           & 71      & 9.22                         & 2.224      
\end{tblr}
\label{table: compare_result}
\end{table}

\begin{table}
\centering
\caption{Processing time breakdown for the proposed method.}
\begin{tblr}{
  column{even} = {c},
  column{3-5} = {c},
  hlines,
  column{3,4}={c}{1.0cm},
  column{5}={c}{.8cm},
  column{2}={c}{1.5cm},
  row{1} = {m},
}
Process   & Equivalent Fluid Flow Problem & Solving via LBM & Sampling  Evaluation &  Total \\
Time (ms) & 28                            & 19              & 36                   & 83       
\end{tblr}
\label{table: processing time}
\end{table}

\subsection{Simulation results}
\label{sec: sim results}
This section provides strong evidence that the motion planner developed in this paper outperforms MPC in terms of efficiency, feasibility and computational cost, while the two algorithms achieve about the same performance concerning safety and comfort. The main reasons for outperforming MPC is that the latter is often trapped around local minima and ideally it requires careful re-tuning of its cost function. In contrast, the proposed method is not sensitive to the selection of the weights of the cost function in Eq.~\eqref{eq: costFunction} because the STVF already provides a near-optimal  solution for the motion planning problem, which is less likely to locate around local minima. Also, recall from the previous section that MPC leverages the best possible manoeuvre in all scenarios, which means that in practice, the proposed motion planner is likely to outperform MPC in all considered KPIs. 

Qualitative and quantitative comparisons are presented in Fig.~\ref{fig: result_compare} and Table~\ref{table: compare_result}, respectively. Fig.~\ref{fig: result_compare} illustrates the control signals as well as the instantaneous speed and acceleration of the EV for scenario \textbf{a1} (motorway overtaking), in which the EV needs to overtake the leading vehicle (OV1), while another vehicle (OV2) approaches from the opposite lane. Since in this scenario, there is not enough time to overtake OV1 before the second vehicle arrives, the EV should gradually reduce its speed and overtake OV1 after OV2 leaves the critical location. Fig.~\ref{fig: result_compare} shows that the proposed motion planner smoothly decelerates down to a minimum speed of $7$~m/s, and changes the lane with low lateral acceleration (maximum absolute of $0.01$~g). MPC brakes with a higher deceleration rate due to the local minimum solution (as if there is no open gap to overtake). The EV reaches the minimum speed of $1.5$~m/s and this leads to a higher steering angle for overtaking. Moreover, MPC generates saturated control signals for the steering angle during this scenario, whereas that is not the case for the proposed method. Nevertheless, the latter generates some jitter in the control signals which should be addressed in future studies. 

The averages of the KPIs over the simulated scenarios ($12$ in total) are presented in Table~\ref{table: compare_result}. In summary, the fluid-based motion planner clearly outperforms MPC in terms of efficiency, feasibility and computational complexity. Concerning comfort and safety, the performances are comparable, however, one should bear in mind that MPC has been simulated for all possible decisions in each scenario, and the performance for the optimal manoeuvre has been finally reported. On the contrary, a key feature of the fluid-based planner is its adaptability in driving situations and ADFs, i.e., it doesn't require specifying the manoeuvre in advance. 

Before closing this section we would like to comment on the time allocation for processing a single time step (excluding initialization) whose breakdown into different tasks is outlined in Table~\ref{table: processing time}. The primary process, LBM, accounts for just 23~\% (19~ms) of the total processing time, with the remaining portion devoted to tasks related to creating the equivalent fluid flow problem and performing sampling procedure. Since this study does not include optimisations or parallelisation for those time-consuming tasks, the overall processing time has the potential to be further reduced, potentially falling below 83~ms. Nevertheless, it's worth noting that this figure remains comfortably below the critical threshold of 100~ms, which is the duration within which up-stream data must be processed.

\section{Summary \& Conclusions}
\label{sec: conclusion}
This study formulates the motion planning problem for automated driving systems (ADS) as a fluid flow problem with boundary conditions. A 3D  spatiotemporal vector field (STVF) is used to sample trajectories, which, according to fluid mechanics principles, are characterised by low energy consumption, hence, improving fuel efficiency. Besides, the fluid-inspired formulation alleviates the need for predefining the manoeuvre, yielding a general-purpose motion planner that can dynamically adapt to diverse driving scenarios, environmental domains and automated driving functions. Performance evaluations in a simulation environment carried out for intersection crossing, motorway overtaking and merging highlight a competitive performance in terms of safety and comfort as compared to state-of-the-art motion planners with model predictive control (MPC). At the same time, the fuel efficiency, the feasibility of control signals (throttle/brake and steeting angle), and the computational cost significantly outperform that of MPC. 
Notably, trajectory generation is only one of many possible applications of fluid dynamics in the realm of motion planning for ADS. Other researchers can explore diverse frameworks to leverage the STVF, e.g., for trajectory prediction of surrounding vehicles. Also, this study can be extended in further studies by comprehensively evaluating the proposed method under realistic conditions by designing a pipeline to test it with publicly available large-scale datasets and comparing results against data-driven approaches.

\appendix
\section{Appendices}
\subsection{Fluid Dynamic Equations of Motion}
\label{app: navier-stokes}
The fluid flow motion is governed by a set of partially differential equations (PDEs) known as the  Navier-Stokes equations~\cite{glowinski2003numerical} that can be read as: 
\begin{align}
    {\epsilon {\bf{U}} - \upsilon \nabla ^2 {\bf{U}} + \nabla p}&= {\bf{f},} &{{\rm{in}}\;\Omega. }  \notag \\
    {\nabla  \cdot {\bf{U}}}&=  {0,} &{{\rm{in}}\;\Omega. } \label{eq: nv} \\
    {\bf{U}} &= {{{\bf{U}}_b},}  &{{\rm{on}}\;\Gamma, } \notag
\end{align}
where ${\bf{U}},{\bf{f}} \in \mathbb{R}{^3}$ are fluid velocity and external force vectors, respectively, $p$ is a scalar value for pressure, $\upsilon$ is the kinematic viscosity (positive value) and $\epsilon$ is a constant. For $\epsilon=0$, Eq.~\ref{eq: nv} becomes the classic Stokes problem and for $\epsilon>0$, it gets in the form of discredited Navier-Stokes equations, which is the base for most of the numerical fluid mechanics simulation methods. Finally, ${{{\bf{U}}_b}}$ is the velocity vector along the boundary $\Gamma$ of the domain $\Omega$.

The Frenet frame is used to construct the fluid motion cuboidal volume (Fig.~\ref{fig: spatiotemporal fluid}-top) from a driving route that might not be a straight line with zero curvature ($\kappa  \ne 0$ in Fig.~\ref{fig: Frenet frame}). Considering this fact, the equivalent fluid flow problem needs compensation for taking non-zero curvatures into account. So, the external force vector in Eq.~\ref{eq: nv} is defined according to the centrifugal and Coriolis accelerations in lateral ($f_d$) and longitudinal ($f_s$) directions respectively:
\begin{equation}
\label{eq: curvature}
{f_d} = \rho \kappa \dot s^2, \quad \;\quad {f_s} = 2\rho \kappa \dot s \dot d,
\end{equation}
where $\rho$ and $\kappa$ are the density and curvature of a point in the simulation volume, and $\dot s$ and $\dot d$ are its longitudinal and lateral velocities, respectively. According to this equation, for the sections of the route where the curvature is zero, the external acceleration field is zero.

\subsection{Lattice Boltzmann Method (LBM)}
\label{app: lbm}
The Lattice Boltzmann Method (LBM) is used as a numerical tool for solving the boundary value problem. This method is explained below in a simplified manner by focusing on 2D problems (D2Q9) for clarity. However, it should be noted that the same methodology can be applied to 3D simulations (D3Q19), as we do in Section~\ref{sec: LBM}.  

\begin{figure}[t!]
\centering
\vspace{3mm}
\includegraphics[width=0.80\linewidth]{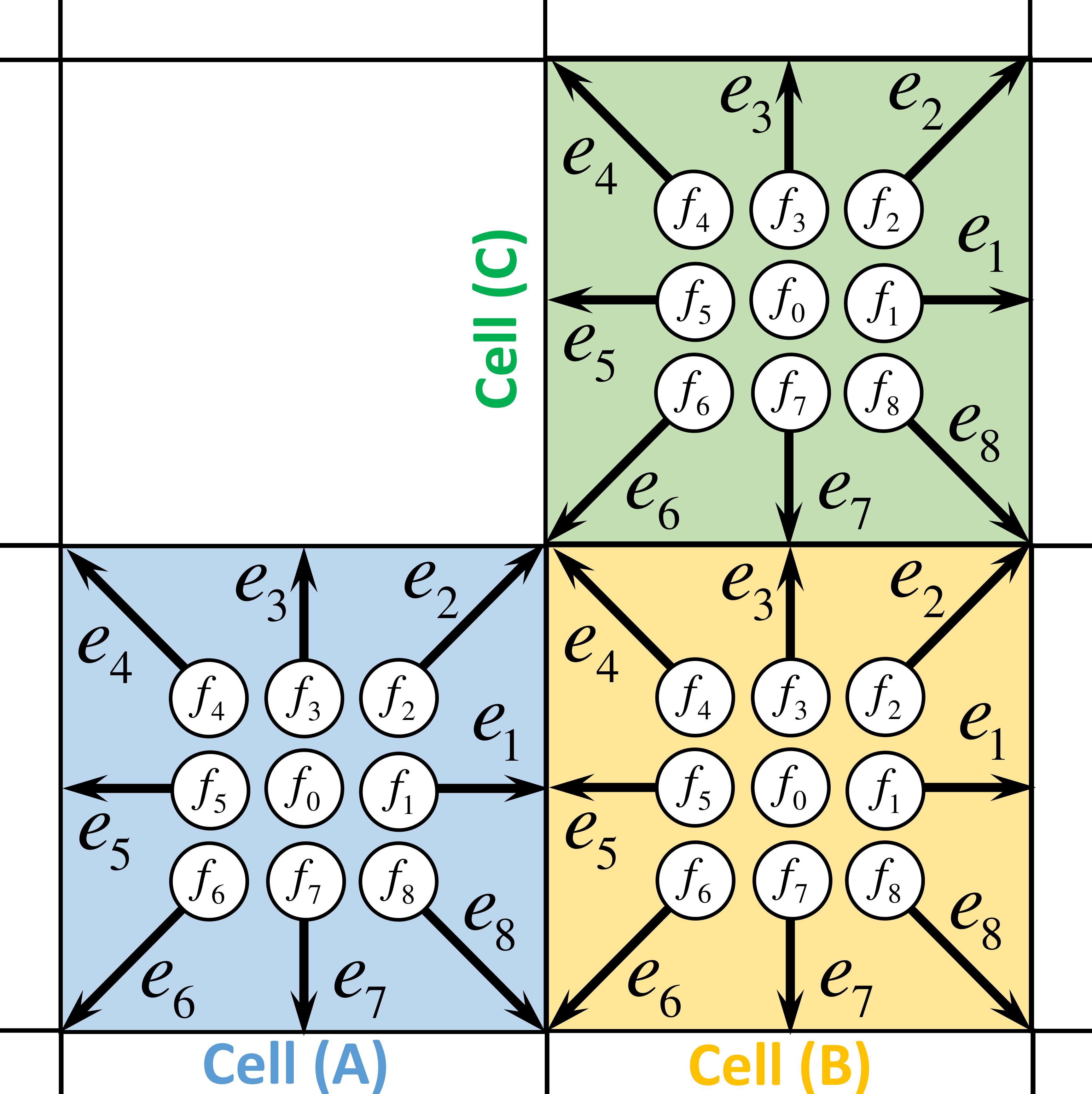}
\caption{Adjacent lattice cells interaction in the overall D2Q9 grid.}
\label{fig: lbm}
\centering
\end{figure}

To generate the 2D velocity vector field using the LBM (D2Q9), each cell in the lattice interacts with eight surrounding cells plus itself (nine directions) via 2D lattice vectors of unit magnitude (${\mathbf{e}_i}$) see Fig.~\ref{fig: lbm}. Firstly, the velocity vectors at the boundary cells are set. Recall from Section~III.A the four boundary conditions used in the fluid flow simulation. Then, the velocity vector (${\bf V}$) for each of the remaining cells is initialized, and based on that a density  $f_i, i=0,\ldots 8$ is assigned to each direction $\mathbf{e}_i$. The values of $(f_i, {\mathbf V})$ for each cell are iteratively updated until there is no significant change in the 
 average velocity ($<~0.01$~m/s) for all cells and the boundary conditions including speed limits are also satisfied. Each iteration consists of the following two steps:

 \textbf{Streaming step:} Adjacent cells exchange the densities between opposite vectors. For instance, in Fig.~\ref{fig: lbm} cell A interacts with cells B and C by exchanging density via $\left\{ {f_5^A \leftarrow f_5^B,f_1^A \to f_1^B} \right\}$ and $\left\{ {f_6^A \leftarrow f_6^C,f_2^A \to f_2^C} \right\}$ pairs, respectively.
 
\textbf{Collision step:} This step is done in each cell separately. First, the equilibrium density is calculated using nine densities and their contributing weights ($\omega_i$). Subsequently, the nine densities and the cell's velocity vector are updated using the equation below (order matters). 
\begin{equation}
\label{eq: collision}
{\begin{array}{*{20}{l}}
{f_i^{eq} = {\omega _i}\rho \left( {1 + 3\, {{{ \mathbf{e}}_i} \bf{V}} - \frac{3}{2} |{\bf{V}}|^2+ \frac{9}{2}{{\left( {{{\mathbf{e}}_i} {\bf{V}}} \right)}^2}} \right)}.\\
{\;\;{f_i} = {f_i} + {{\left( {f_i^{eq} - {f_i}} \right)} \mathord{\left/
 {\vphantom {{\left( {f_i^{eq} - {f_i}} \right)} \tau }} \right.
 \kern-\nulldelimiterspace} \tau_0. }}\\
{\;\;\,\,\rho  = \sum\limits_{i = 0}^8 {{f_i}}. }\\
{\;\;\,{\bf{V}} = \sum\limits_{i = 0}^8 {{f_i}\, {{\mathbf{e}}_i}} }.
\end{array}}
\end{equation}

According to~\cite{januszewski2014sailfish}, the updating weights for stationary (${\bf e}_0$), diagonal ($\mathbf{e}_i, i\in \{2,4,6,8\}$), and orthogonal ($\mathbf{e}_i, i\in \{1,3,5,7\}$) directions are $\frac{4}{{9}}$, $\frac{1}{{36}}$ and $\frac{1}{{9}}$, respectively. Finally, in fluid flow simulations, $\tau_0$ is the update rate obtained from the kinematic viscosity property of the fluid which has been set equal to $0.003$ in this study. It should be noted that at each update iteration, the velocity vectors corresponding to the boundary conditions are set to the boundary values.

\subsection{Forward and Inverse Dynamics}
\label{app: inverse dynamic}
The forward dynamics calculate the rate of change in the state of the vehicle given the control inputs, whereas, the inverse dynamics obtain the required inputs to achieve a desired rate of change in the current state of the vehicle. Forward dynamics simply evaluate Eq.~\eqref{eq: dynamic eq} given the force $F_x$ and steering angle $\delta_f$. In contrast, due to the nonholonomic constraints in the dynamic model, the calculation of the inverse dynamics is not trivial. The input to the inverse dynamics function  $\mathcal{F}^{-1}$ are the current state of the EV (${\rm E}$) and the lateral/longitudinal accelerations in the vehicle frame (${\dot u},{\dot v}$). The input signals are calculated via two main steps according to Algorithm~\ref{alg: InvDyn}. First, calculating the throttle/brake force $F_x$, steering angle $\delta_f$, and the derivative of yaw rate ($\dot r$) from Eq.~\eqref{eq: dynaimc eq1} to \eqref{eq: dynaimc eq3} (\textbf{lines 1-3}). Second, calculating the upper and lower bound of the yaw rate  (\textbf{lines 4, 5}), where the tire slip angle limits depends on the tire model and are set to $\{ {\alpha _{\min }},{\alpha _{\max }}\}  = \{  - 4,4\}$ degrees~\cite{gao2014tube}. 
In order to ensure staying within the linear tire slip angle limits $r\in (r_{\min}, r_{\max})$, if the calculated value of the yaw rate ($r$) cause out-of-range slip angles and the magnitude of yaw rate increases (\textbf{lines 6-10}), $\dot r$ is set to zero and the steering angle is updated accordingly.

\SetCommentSty{xCommentSty}
\setlength{\algomargin}{0pt}

\RestyleAlgo{ruled}
\SetKwComment{Comment}{/* }{ */}
\SetKw{KwBy}{by}
\SetKwInOut{Input}{Input}
\SetKwInOut{Output}{Output}
\IncMargin{1.3em}
\begin{algorithm}[t]
\caption{Inverse dynamics function $\mathcal{F}^{-1}(\cdot)$}
\label{alg: InvDyn}
\DontPrintSemicolon
\Input{%
${\rm E}$ -- Current state of EV\\
$(\dot u,\dot v)$ -- Desired accelerations\\
} 

\Output{$(F_x,\delta_f)$ -- Required input signals\\}

\BlankLine
$F_x \gets$ Substitute (${\rm E}$, $\dot u$, $\dot v$) in Eq.~\eqref{eq: dynaimc eq1}\\
$\delta_f \gets$ Substitute (${\rm E}$, $\dot u$, $\dot v$) in Eq.~\eqref{eq: dynaimc eq2}\\
$\dot r \gets$ Substitute (${\rm E}$, $\delta_f$) in Eq.~\eqref{eq: dynaimc eq3}\\

\Comment*[r]{linear region (see~\cite{gao2014tube}):} 
$r_{max} \gets$ $\min \left\{ {{{\left( {\left( {{\alpha _{\max }} + {\delta _f}} \right)u - v} \right)} \mathord{\left/
 {\vphantom {{\left[ {v + \left( {{\alpha _{\max }} - {\delta _f}} \right)u} \right]} {{l_f}}}} \right.
 \kern-\nulldelimiterspace} {{l_f}}},{{\left( {v - {\alpha _{\max }}u} \right)} \mathord{\left/
 {\vphantom {{\left( {v + {\alpha _{\max }}u} \right)} {{l_r}}}} \right.
 \kern-\nulldelimiterspace} {{l_r}}}} \right\}$ \\
 $r_{min} \gets$ $\max \left\{ {{{\left( {\left( {{\alpha _{\min }} + {\delta _f}} \right)u -v} \right)} \mathord{\left/
 {\vphantom {{\left[ {v + \left( {{\alpha _{\min }} - {\delta _f}} \right)u} \right]} {{l_f}}}} \right.
 \kern-\nulldelimiterspace} {{l_f}}},{{\left( {v - {\alpha _{\min }}u} \right)} \mathord{\left/
 {\vphantom {{\left( {v + {\alpha _{\min }}u} \right)} {{l_r}}}} \right.
 \kern-\nulldelimiterspace} {{l_r}}}} \right\}$ \\
\If{$\left( {{r } < {r _{\min }}\;{\bf{OR}}\;{r } > {r _{\max }}} \right)\;{\bf{AND}}\;r\dot r < 0$}{
 $\dot r \gets 0$\\
  $\delta_f \gets$ Substitute (${\rm E}$, $\dot r$) in Eq.~\ref{eq: dynaimc eq3}\\
    }
\Return{($F_x,\delta_f$)}
\end{algorithm}

\section*{Acknowledgement}
This research is sponsored by the Centre for Doctoral Training to Advance the Deployment of Future Mobility Technologies at the University of Warwick. This work is also part of the Hi-Drive project. The project has
received funding from the European Union’s Horizon 2020 research
and innovation programme under grant agreement No. 101006664.

\bibliographystyle{myIEEEtran}
\bibliography{IEEEabrv,bibliography}

\vskip 0pt plus -1fil

\begin{IEEEbiography}[{\includegraphics[width=1in,height=1.25in,clip,keepaspectratio]{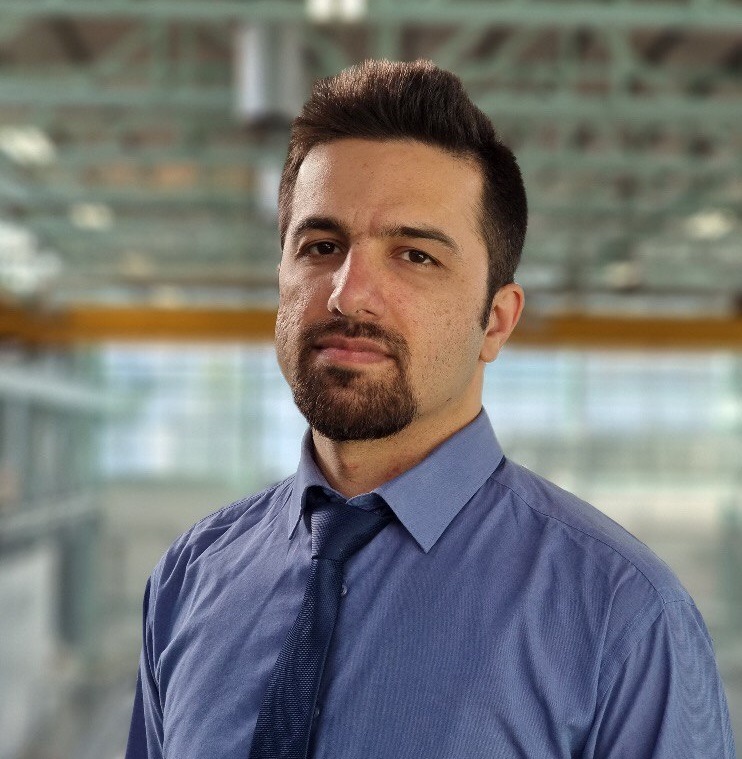}}]{Mohammadreza Alipour Sormoli}
received the M.Sc. degree from the Amirkabir University of Technology (Tehran Polytechnic) in 2017. worked as a research assistant at Koc University and is currently working toward the PhD degree in the field of autonomous driving technology at the University of Warwick (WMG). His research interests include robotics, mechatronics, control and dynamics of autonomous systems.
\end{IEEEbiography}

\begin{IEEEbiography}[{\includegraphics[width=1in,height=1.25in,clip,keepaspectratio]{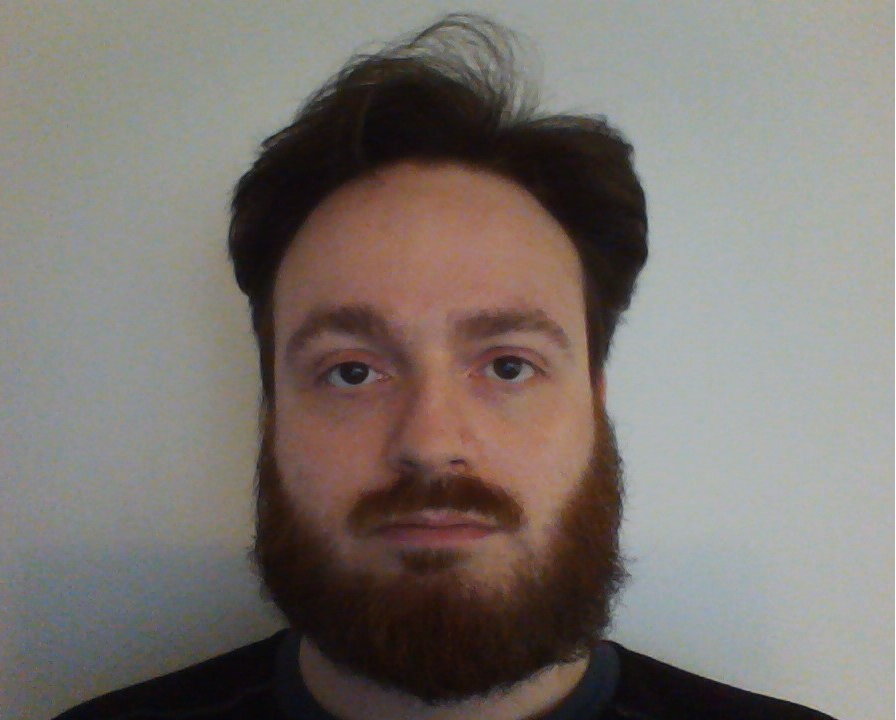}}]{Konstaninos Koufos} obtained the 5-year Diploma in Electrical \& Computer Engineering from Aristotle University of Thessaloniki, Greece, in 2003 and the M.Sc. and D.Sc. in Radio Communications from Aalto University, Finland, in 2007 and 2013. He worked as a post-doctoral researcher on 5G wireless networks at Aalto University, as a Senior Research Associate in Spatially Embedded Networks at the School of Mathematics at the University of Bristol, UK, and as a Senior Research Fellow on Cooperative Autonomy within the Warwick Manufacturing Group (WMG) in the University of Warwick, UK. He is an Assistant Professor on Future Mobility Technology at WMG, working on perception, motion planning, and fail-safe control of connected and automated vehicles. His research interests also include stochastic geometry and performance modelling of mobile wireless networks.
\end{IEEEbiography}

\begin{IEEEbiography}[{\includegraphics[width=1in,height=1.25in,clip,keepaspectratio]{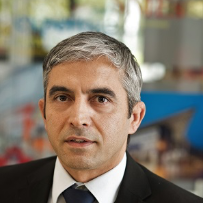}}]{Mehrdad Dianati}
(Senior Member, IEEE) is a professor of connected and cooperative autonomous vehicles at WMG, the University of Warwick and the School of EEECS at the Queen's University of Belfast. He has been involved in a number of national and international projects as the project leader and the work-package leader in recent years. Prior to academia, he worked in the industry for more than nine years as a Senior Software/Hardware Developer and the Director of Research and Development. He frequently provides voluntary services to the research community in various editorial roles; for example, he has served as an Associate Editor for the IEEE Transactions On Vehicular Technology.  He is the Field Chief Editor of Frontiers in Future Transportation.
\end{IEEEbiography}

\begin{IEEEbiography}[{\includegraphics[width=1in,height=1.25in,clip,keepaspectratio]{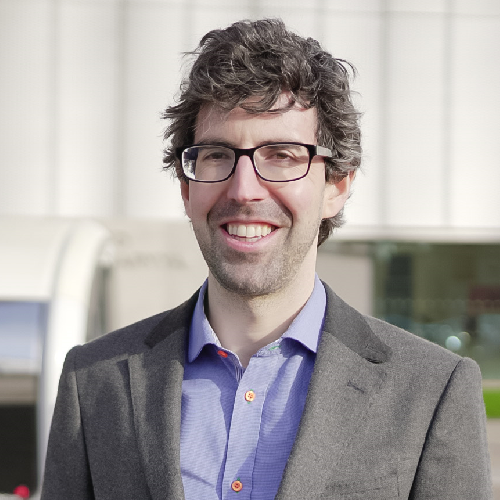}}]{Roger Woodman}
is an Assistant Professor and Human Factors research lead at WMG, University of Warwick. He received his PhD from Bristol Robotics Laboratory and has more than 20 years of experience working in industry and academia. Among his research interests, are trust and acceptance of new technology with a focus on self-driving vehicles, shared mobility, and human-machine interfaces. He has several scientific papers published in the field of connected and autonomous vehicles. He lectures on the topic of Human Factors of Future Mobility and is the Co-director of the Centre for Doctoral Training, training doctoral researchers in the areas of intelligent and electrified mobility systems.
\end{IEEEbiography}

\end{document}